%% file: main.tex
\documentclass[final, 3p, 12pt, times]{elsarticle}

\usepackage{amssymb}
\usepackage{amsmath}

\newtheorem{definition}{Definition}
\usepackage{multirow}
\usepackage{array}
\usepackage{subcaption}
\usepackage{hyperref}
\usepackage{xcolor}
\usepackage{booktabs}

\begin{document}

\begin{frontmatter}



\title{Hypergraph-based Motion Generation with Multi-modal Interaction Relational Reasoning}


\author[label1,label2]{Keshu Wu} 
\author[label2]{Yang Zhou\corref{cor1}} 
\author[label3]{Haotian Shi\corref{cor1}} 
\author[label2]{Dominique Lord}
\author[label3]{Bin Ran} 
\author[label1]{Xinyue Ye}
\affiliation[label1]{organization={Center for Geospatial Sciences, Applications, and Technology and Department of Landscape of Architecture and Urban Planning, Texas A\&M University},
            addressline={788 Ross St}, 
            city={College Station},
            postcode={77840}, 
            state={TX},
            country={United States}}
\affiliation[label2]{organization={Zachry Department of Civil and Environmental Engineering, Texas A\&M University},
            addressline={201 Dwight Look Engineering Building}, 
            city={College Station},
            postcode={77843}, 
            state={TX},
            country={United States}}
\affiliation[label3]{organization={Department of Civil and Environmental Engineering, University of Wisconsin-Madison},
            addressline={1415 Engineering Dr}, 
            city={Madison},
            postcode={53706}, 
            state={WI},
            country={United States}}

\cortext[cor1]{\raggedright Corresponding authors:Yang Zhou (\href{mailto:yangzhou295@tamu.edu}{yangzhou295@tamu.edu}), Haotian Shi (\href{mailto:hshi84@wisc.edu}{hshi84@wisc.edu}); }
 
\begin{abstract}
The intricate nature of real-world driving environments, characterized by dynamic and diverse interactions among multiple vehicles and their possible future states, presents considerable challenges in accurately predicting the motion states of vehicles and handling the uncertainty inherent in the predictions. Addressing these challenges requires comprehensive modeling and reasoning to capture the implicit relations among vehicles and the corresponding diverse behaviors. This research introduces an integrated framework for autonomous vehicles (AVs) motion prediction to address these complexities, utilizing a novel \textbf{R}elational \textbf{H}ypergraph \textbf{I}nteraction-informed \textbf{N}eural m\textbf{O}tion generator (\texttt{RHINO}). \texttt{RHINO} leverages hypergraph-based relational reasoning by integrating a multi-scale hypergraph neural network to model group-wise interactions among multiple vehicles and their multi-modal driving behaviors, thereby enhancing motion prediction accuracy and reliability. Experimental validation using real-world datasets demonstrates the superior performance of this framework in improving predictive accuracy and fostering socially aware automated driving in dynamic traffic scenarios. The source code is publicly available at \url{https://github.com/keshuw95/RHINO-Hypergraph-Motion-Generation}.
\end{abstract}




\end{frontmatter}



\section{Introduction}

\begin{figure}[!ht]
\centering
\includegraphics[width=0.8\textwidth]{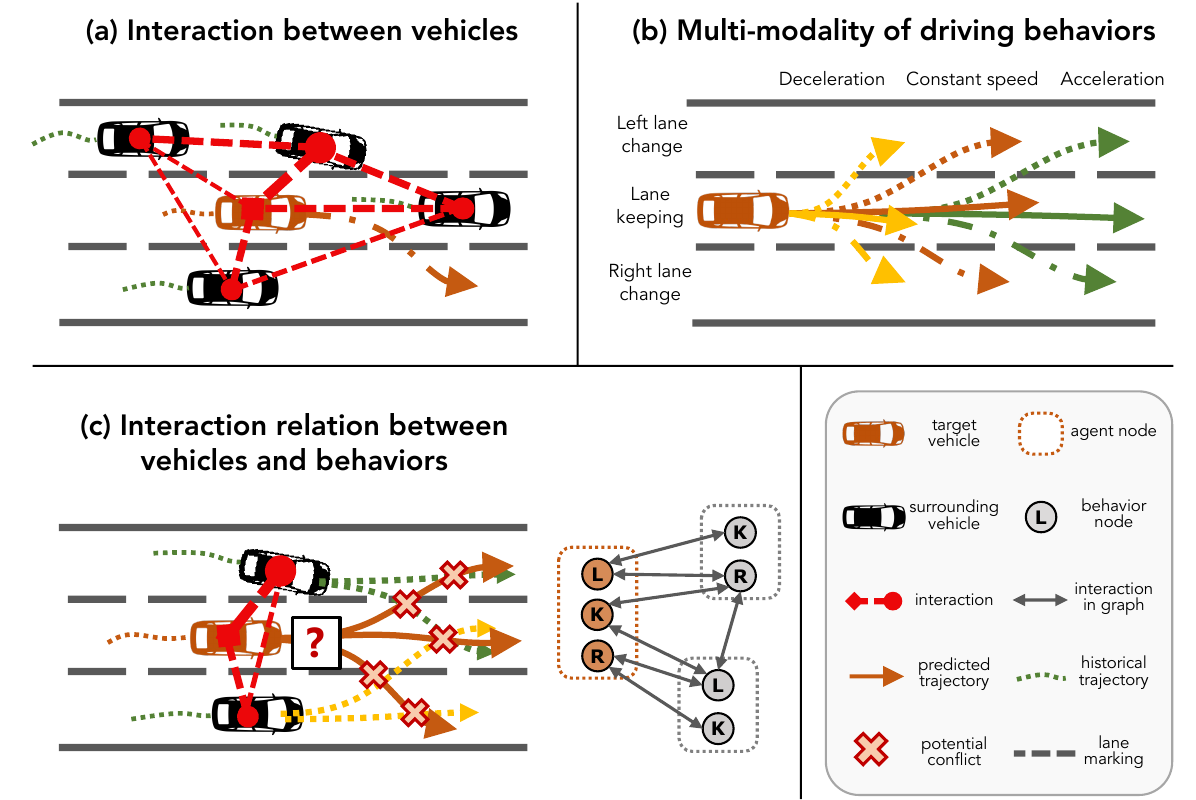}
\caption{Major challenges: (a) vehicle interaction, (b) behavior multi-modality, and (c) interaction relational reasoning.}
\label{fig_challenges}
\end{figure}

Understanding traffic interactions and the way they affect future vehicle trajectories is inherently complex \cite{gao2020vectornet}. In mixed traffic environments, where human-driven and automated vehicles coexist, this complexity is amplified, requiring precise interaction representation and behavior modeling for reliable motion prediction \cite{li2020synchronous, shi2021connected, shi2023physics, wang2025generative}. These scenarios often present dynamic interaction topologies with underlying relations, with interaction patterns as well topologies continuously evolving depending on the surrounding context as exampled by lane change maneuvers \cite{zhou2024reasoning, liu2023predictive}. These relationships play a crucial role in guiding each vehicle's decision-making processes. Additionally, each vehicle can display multiple possible modalities in driving intentions and behaviors, including both longitudinal (e.g., acceleration and braking) and lateral (e.g., lane-changing and lane-keeping) maneuvers \cite{suo2021trafficsim, deo2018multi, you2024followgen}. Furthermore, collective behaviors, arising from interactions within a group of vehicles and encompassing both cooperative and competitive behaviors \cite{wang2021competitive, trentin2023multi, karle2022scenario, ali2020cooperate}, further complicate the understanding of these interactions\cite{li2020generative, han2024collective, zhong2024understanding}. Figure \ref{fig_challenges} describes the interaction among multi-vehicles and the corresponding multi-modal driving behaviors. Therefore, it is necessary to model the interaction and multi-modality and reason the interaction relation to accurately capture interactions and forecast their future behaviors \cite{kipf2018neural, li2020evolvegraph, xu2024dynamic}.
 
Efforts have been made to address the challenges of vehicle interactions and driving behavior multi-modality. Three primary approaches have been developed: social operation methods, attention-driven methods, and graph-based techniques. Social operations use pooling mechanisms to generate socially acceptable trajectories by capturing the influence of surrounding agents \cite{alahi2016social, gupta2018social}. Attention-driven approaches use attention mechanisms to dynamically weigh neighboring agents' information \cite{chen2022intention, zhang2022ai, kim2020multi, gan2025goal}. Graph-based methods leverage graph structures to model non-Euclidean spatial dependencies, effectively handling varying interaction topologies and predicting dynamic interactions \cite{wu2023graph, li2019grip, zhou2021ast, huang2019stgat}. These complex interactions create uncertainty, complicating the accurate forecasting of a single future trajectory with high confidence due to varying driving behaviors in identical situations \cite{amirian2019social, khandelwal2020if}, driven by individual driver characteristics and psychological factors. Addressing the multi-modality of driving behaviors often involves introducing latent variables, categorized into those with explicit semantics and those without. Models with explicit semantics use latent variables to clearly represent driving intentions, identifying specific maneuvers and behaviors for multi-modal trajectory predictions \cite{khandelwal2020if, deo2018convolutional, messaoud2020attention}. Conversely, models without explicit semantics employ generative deep learning techniques, such as Variational Autoencoders (VAEs) \cite{feng2019vehicle} and Generative Adversarial Networks (GANs) \cite{gupta2018social, wang2020multi}, to produce diverse trajectories by adding noise to encoded features. While these models generate a wide range of possible trajectories, they often struggle with issues related to interpretability and identifying the most effective strategies for sampling from the generated trajectories. 

\begin{figure}[!ht]
\centering
\includegraphics[width=0.75\textwidth]{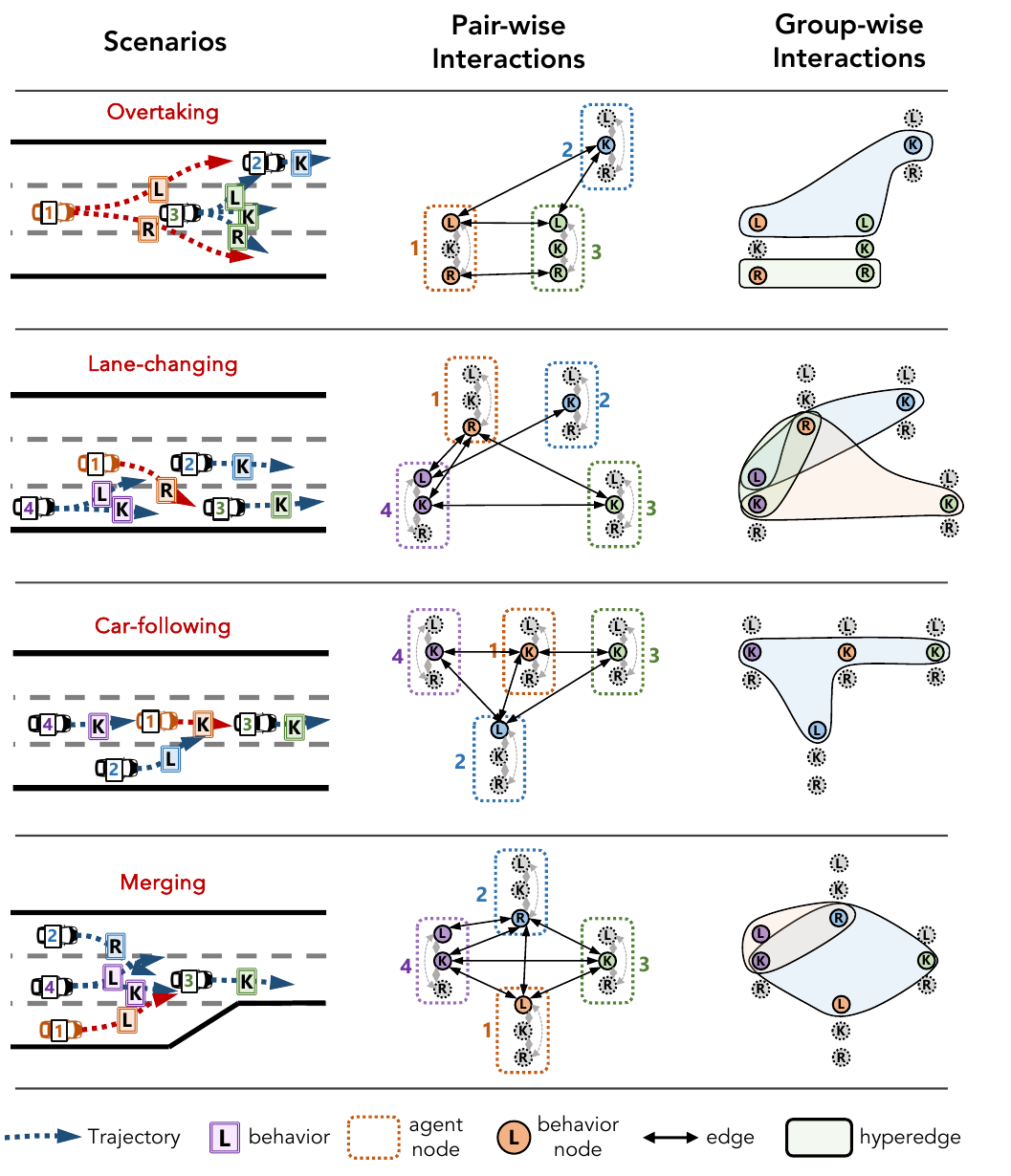}
\caption{Pair-wise interaction and group-wise interaction in different scenarios.}
\label{fig_group_wise_interaction}
\end{figure}

Despite the advances in modeling vehicle interactions and probabilistically forecasting multi-modal future trajectories, the inherent complexity of social dynamics in traffic systems continues to present significant challenges, with limitations arising from the complex nature of agent interactions in two key aspects: First, most predictors entangle all plausible maneuvers in a single node and restrict influence to pairwise edges, which blurs intention and weakens interaction reasoning. Prevailing methods primarily use a single graph where each vehicle is represented by one node and edges encode only binary, pairwise interactions. This conflates distinct behavior modes within a single embedding and aggregates messages without conditioning on the maneuver each agent is likely to execute. In multi-vehicle systems, however, interactions are frequently cooperative or competitive \cite{wang2021competitive, trentin2023multi, zhou2023data, huang2021driving, zhou2024deep} and depend on which maneuvers neighboring agents pursue. When behaviors are merged, the model tends to average over incompatible futures, obscuring intention and degrading prediction. For instance, in Fig.~\ref{fig_group_wise_interaction}, vehicle 1 attempts a right lane change (R), vehicle 3 keeps lane (K), and vehicle 4 executes a left lane change (L), while platooning emerges when vehicles 1, 3, and 4 maintain lane keeping in response to vehicle 2’s left lane change. These cases motivate a representation that disentangles behavior at the node level so that interaction reasoning can be carried out per mode, rather than over a single, behavior-averaged embedding.

Second, binary-edge graphs, even with attention, cannot natively encode higher-order, many-to-many influences; hyperedges provide the missing interaction substrate. While traditional graph-based methods such as \cite{wu2023graph, li2019grip, zhou2021ast, sheng2022graph, sheng2024kinematics, sheng2024ego} excel at capturing pair-wise relationships, they are limited in representing the more intricate group-wise interactions. This limitation arises because graph-based approaches model interactions between pairs of agents with fixed topologies, making it difficult to capture the simultaneous influence of multiple agents on each other’s behaviors. By contrast, a hypergraph generalizes a graph by allowing a single hyperedge to connect an entire group of nodes within a shared context \cite{gao2020hypergraph, feng2019hypergraph, wu2025ai2, wang2025nest}. This explicit many‐to‐many structure more accurately captures group-wise behaviors and manages the uncertainty and variability arising from these group interactions. This adaptability is essential for addressing the stochastic nature of human behavior, as hypergraph models can more effectively manage the uncertainty and variability inherent in the interactions resulting from collective behaviors, thereby facilitating more socially inspired automated driving \cite{zhou2022grouptron, xu2022groupnet, li2024multi, li2025simulating}.

To address the aforementioned challenges, this paper presents a novel hypergraph-based method for multi-modal trajectories prediction with relational reasoning. The proposed framework contains two parts: \textbf{G}raph-based \textbf{I}nteraction-awa\textbf{R}e \textbf{A}nticipative \textbf{F}easible \textbf{F}uture \textbf{E}stimator (\texttt{GIRAFFE}) and \textbf{R}elational \textbf{H}ypergraph \textbf{I}nteraction-informed \textbf{N}eural m\textbf{O}tion generator (\texttt{RHINO}). \texttt{GIRAFFE} enables multi-agent, multi-modal motion prediction of preliminary multi-vehicle trajectories, based on while \texttt{RHINO} framework, which utilizes an innovative Agent-Behavior Hypergraph to capture group-wise interactions among various behavior modalities and motion states. Leveraging \texttt{GroupNet} \cite{xu2022groupnet} as its backbone, \texttt{RHINO} learns a multi-scale hypergraph topology in a data-driven manner to model group-wise interactions. Through neural message passing across the hypergraph, this approach integrates interaction representation learning and relational reasoning, enhancing the social dynamics of automated driving. Furthermore, a Conditional Variational Autoencoder (CVAE) framework is employed to generate diverse trajectory predictions by sampling from hidden feature distributions, effectively addressing the stochastic nature of agent behaviors.

To summarize, the key contributions of this work are as follows:

\begin{enumerate} 
    \item The framework adopts multi-scale hypergraphs to represent group-wise interactions among different modalities of driving behavior and the corresponding motion states of multiple agents in a flexible manner.
    
    \item This framework incorporates interaction representation learning and relational reasoning to generate motions that are plausible and concurrently in a probabilistic manner depicted by the learned posterior distribution.
\end{enumerate}

The remainder of this paper is structured as follows: Section 2 outlines the problem statement of this research. Section 3 introduces the methodology. Section 4 details the experimental setup and analysis of the results obtained. Section 5 concludes this paper.

\section{Problem Statement}
\subsection{Problem Definition}
The vehicle trajectory prediction in the dynamic realm of multi-vehicle interaction context of multi-lane highways involves determining the future movements of a target vehicle based on historical data and multi-modal predictions of its own state and the states of surrounding vehicles. This domain addresses two primary challenges: (i) multi-agent multi-modal trajectory prediction and (ii) prediction-guided motion generation after reasoning.

The objective of trajectory prediction is to estimate the future trajectories of the target vehicle and its surrounding vehicles, given their historical states. The historical states, spanning a time horizon $[1, \dots, T]$, are represented as $\mathbf{X}_{1:T} = \{\mathbf{X}_1, \mathbf{X}_2, \dots, \mathbf{X}_T\} \in \mathbb{R}^{T \times N \times C_1}$, where $N$ denotes the number of vehicles and $C_1$ denotes the number of features, including longitudinal and lateral positions and velocities. Each historical state $\mathbf{X}_t = \{ \mathbf{x}_t^i | \forall i \in [1, N], \forall t \in [1, T]\} \in \mathbb{R}^{C_1}$ at time step $t$ captures these details for each vehicle $i$. Notably, the superscript refers to vehicle indices with $i=1$ representing the target vehicle, and the subscript to time steps, with $C_1=4$ for the input data. 

The prediction model, $\mathbf{H}^{Pred}(\cdot)$, provides preliminary predictions of multi-modal trajectory candidates $\hat{\mathbf{X}}_{T+1:T+F}^M \in \mathbb{R}^{F \times N \times M \times C_2}$ for all the $N$ vehicles over the future time horizon $[T+1, \dots, T+F]$ with $M$ modes of driving behaviors. This model takes historical data $\mathbf{X}_{1:T}$ as the input and outputs future longitudinal and lateral positions, where $C_2=2$. The forecasted states $\hat{\mathbf{X}}_f^M = \{\mathbf{x}_f^{m,i} | \forall m \in [1, M], \forall i \in [1, N], \forall f \in [1, F]\}$ aim to estimate each vehicle's future trajectory for each behavior mode $m$ at time step $T+f$. This is mathematically formulated as:
\begin{equation}
\hat{\mathbf{X}}_{T+1:T+F}^M = \mathbf{H}^{Pred}(\mathbf{X}_{1:T})
\end{equation}

Based on that, the motion generation model $\mathbf{H}^{Gen}(\cdot)$ is further developed to generate plausible trajectories considering the implicit group-wise interactions, using both historical states $\mathbf{X}_{1:T}$ and preliminary multi-modal future trajectory candidates $\hat{\mathbf{X}}_{T+1:T+F}^M$ as the input. The generation model provides $K$ plausible trajectory $\hat{\mathbf{Y}}_{T+1:T+F}^K = \{\hat{\mathbf{Y}}_{T+1}^K, \dots, \hat{\mathbf{Y}}_{T+F}^K\} \in \mathbb{R}^{F \times N \times K \times  C_2}$ for all the $N$ vehicles for the next $F$ time steps. Each generated state $\hat{Y}_{T+f}^K = \{\hat{\mathbf{y}}_{T+f}^{k,i} | \forall k \in [1, K], \forall i \in [1, N], \forall f \in [1, F]\} \in \mathbb{R}^{C_2}$ represents the $k$-th generated longitudinal and lateral potisions of the $i$-th vehicle at time step $T+f$. The formulation for this generation problem is:
\begin{equation}
\hat{\mathbf{Y}}_{T+1:T+F}^K = \mathbf{H}^{Gen}(\mathbf{X}_{1:T}, \hat{\mathbf{X}}_{T+1:T+F}^M)
\end{equation}

\section{Methodology}
Given the aforementioned problem, we first develop a customized framework architecture. Then, the vital components are further elaborated.

\subsection{Framework Architecture}
\begin{figure}[!ht]
\centering
\includegraphics[width=1\textwidth]{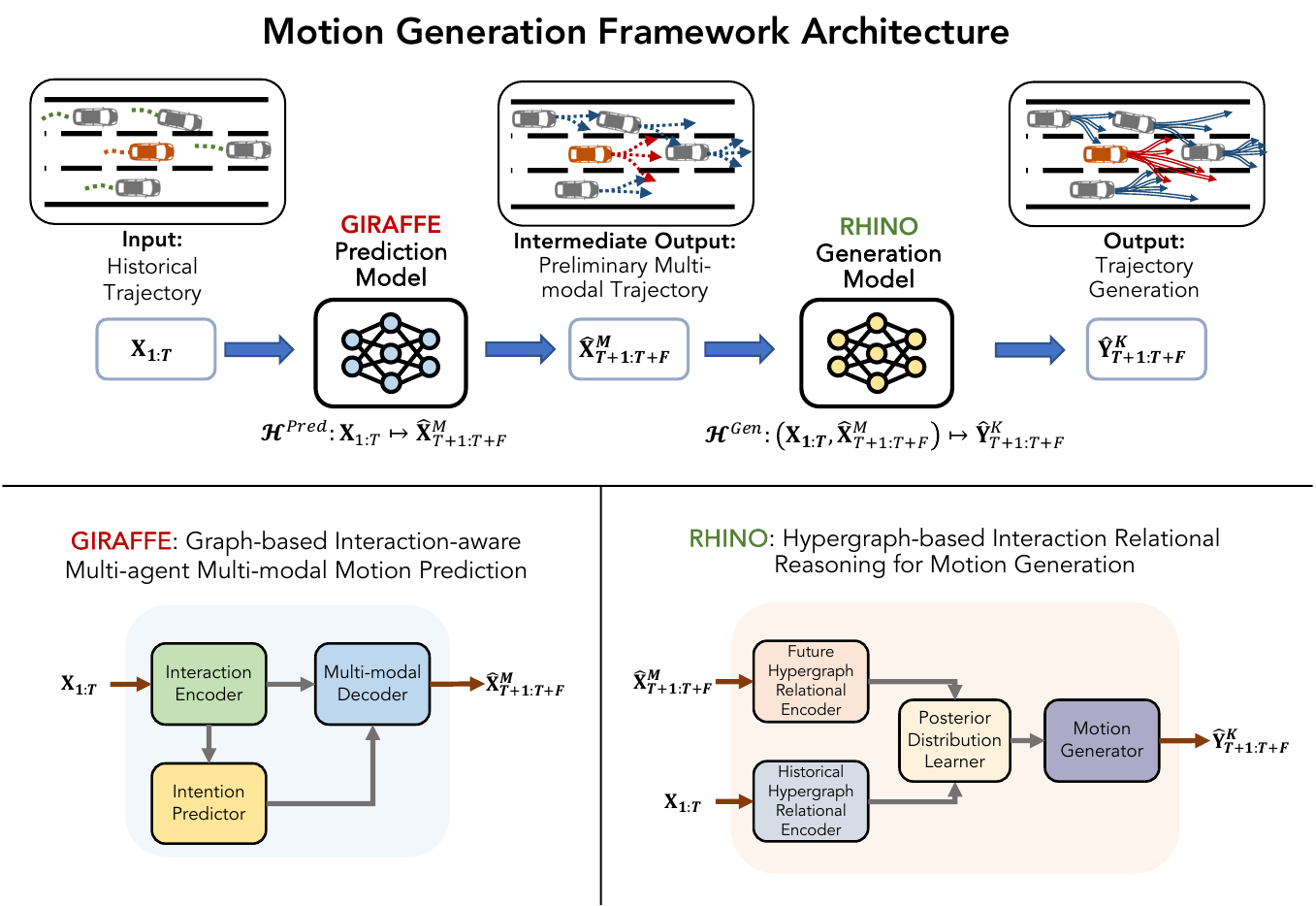}
\caption{Framework architecture.}
\label{fig_framework}
\end{figure}

The proposed framework adopts an integrated architecture, as shown in Figure \ref{fig_framework}, which involves two major components: 
\begin{itemize}
    \item \texttt{GIRAFFE}: \textbf{G}raph-based \textbf{I}nteraction-awa\textbf{R}e \textbf{A}nticipative \textbf{F}easible \textbf{F}uture \textbf{E}stimator, which leverages graph representations to capture pair-wise interactions during both the historical and future time horizons, providing preliminary multi-modal trajectories prediction candidates for vehicles.
    \item \texttt{RHINO}: \textbf{R}elational \textbf{H}ypergraph \textbf{I}nteraction-informed \textbf{N}eural m\textbf{O}tion generator, which utilizes multi-scale hypergraph representations to model group-wise interactions and reason the interaction relations among the multi-modal behaviors. Built upon the preliminary multi-modal trajectories by \texttt{GIRAFFE} , \texttt{RHINO} will further generate plausible future trajectories for all vehicles in a probabilistic manner.
\end{itemize}

The subsequent sections will provide an in-depth explanation of the two principal frameworks.

\subsection{GIRAFFE: Graph-based Motion Predictor}
In our context, a graph representation $\mathcal{G}$ is adopted by modeling $N$ vehicles as nodes $\mathcal{V} \in \mathbb{R}^N$ and the pair-wise interaction as the edges $\mathcal{E} \in \mathbb{R}^{|N\times N|}$. Further, The feature matrix $X \in \mathbb{R}^{N \times C}$ containing vehicle states (i.e., longitudinal and lateral position and speed) and the adjacency matrix $A \in \mathbb{R}^{N \times N}$ describing the interactions among nodes are further utilized to describe the graph. By that, we can define an Agent Graph as:
\begin{definition} [Agent Graph]
    Let $\mathcal{G}^a$ be a graph representing the motion states and interaction of $N$ agents, with each agent represented as a node. $\mathcal{G}^a$ is expressed as 
    $$\mathcal{G}^a=(\mathcal{V}^a, \mathcal{E}^a; X^a, A^a)$$ 
    where $\mathcal{V}^a \in \mathbb{R}^N$ denotes the node set, $\mathcal{E}^a \in \mathbb{R}^{|N\times N|}$ denotes the edge set, $X^a \in \mathbb{R}^{N \times C}$ represents the feature tensor, $A^a \in \mathbb{R}^{N \times N}$ indicates the adjacency matrix.
\end{definition}

To better represent the interaction and relations of the predicted multi-agent multi-modal trajectory candidates with graphs, we expand each agent node to multiple nodes of the number of behavior modes based on our previous work \cite{wu2023graph}, which further renders an Agent-Behavior Graph.

\begin{definition} [Agent-Behavior Graph]
    Let $\mathcal{G}^b$ be a graph representation of the multi-modal motion states of $N$ agents, with each of $M$ behavior modes for each agent represented as a node. $\mathcal{G}^b$ is expressed as
    $$\mathcal{G}^b=(\mathcal{V}^b, \mathcal{E}^b; X^b, A^b)$$ 
    where $\mathcal{V}^b \in \mathbb{R}^{|MN|}$ denotes the node set, $\mathcal{E}^b \in \mathbb{R}^{|MN \times MN|}$ denotes the edge set, $X^b \in \mathbb{R}^{|MN| \times C}$ represents the feature tensor, $A^b \in \mathbb{R}^{|MN| \times |MN|}$ indicates the adjacency matrix.
\end{definition}

\paragraph{\textbf{Behavior‐Mode Definition}}
To explicitly enumerate each vehicle’s possible maneuvers, we decompose behavior into three lateral modes:
\begin{equation} 
  \mathcal{M}_{\rm lat}
    = \{\,L\,(\text{Left lane change}),\;
             K\,(\text{Lane keeping}),\;
             R\,(\text{Right lane change})\},
  \quad M = 3.
\end{equation}
Each agent $i$ thus expands into three nodes
$\{v^b_{i,L},\,v^b_{i,K},\,v^b_{i,R}\}\subset\mathcal{V}^b$.  By masking out any infeasible mode at each time step (e.g.\ $L$ in the leftmost lane), we ensure only valid lane‐change hypotheses enter the graph. The detailed definition of the longitudinal and lateral maneuvers can be found in \ref{app_behavior_mode}.

\paragraph{\textbf{From Agent Graph to Agent-Behavior Graph}}
The transition from an Agent Graph $\mathcal{G}^a$ to an Agent-Behavior Graph $\mathcal{G}^b$ is by an expansion function $\mathbf{F}^{expand}(\cdot)$ as:

\begin{equation} 
    \mathcal{G}^{b} = \mathbf{F}^{expand} \left( \mathcal{G}^a \right)
    \Leftrightarrow 
    \left\{ 
    \begin{aligned} 
        \mathcal{V}^b &= \bigcup_{i=1}^{N} \bigcup_{m=1}^{M}  \left\{ v_{i,m}^b \right\} \\ 
        \mathcal{E}^b &= 
        \bigcup_{i=1}^{N} \bigcup_{j=1}^{N} 
        \left\{ e_{ij,mn}^b \mid A^a_{ij} \neq 0 \text{ and } \Lambda_{mn} \neq 0, \forall m,n \in \left\{1,\dots, M \right\}\right\} \\
        X^b &= \bigcup_{i=1}^{N} \bigcup_{m=1}^{M} X_{i,m}^a \\ 
        A^b &= \bigcup_{i=1}^{N} \bigcup_{j=1}^{N} \left\{ A^a_{ij} \otimes \Lambda_{mn} \mid \forall m,n \in \left\{1,\dots, M \right\} \right\}
        \end{aligned} 
    \right.
\end{equation}

\begin{figure}[!ht]
\centering
\includegraphics[width=0.9\textwidth]{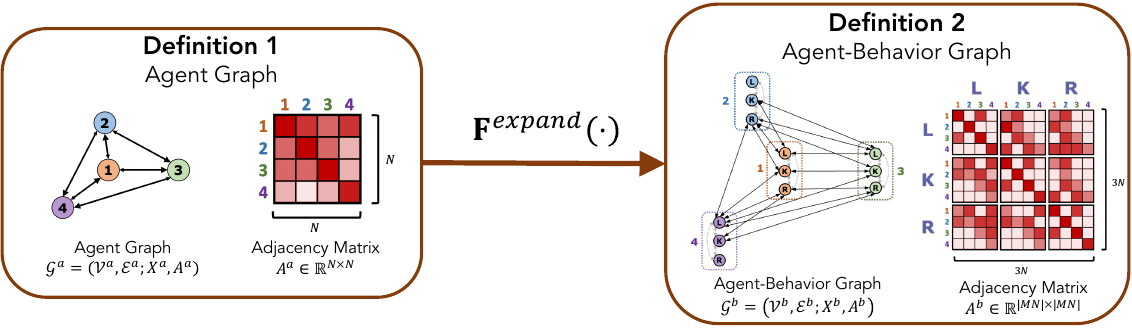}
\caption{Definitions of graphs}
\label{fig_graph_def}
\end{figure}

As shown in Figure \ref{fig_graph_def}, in this process, each agent node $v_i^a \in \mathcal{V}^a$ in the Agent Graph is expanded into $M$ behavior-specific nodes $v_i^b \in \mathcal{V}^b$, corresponding to the $M$ potential behavioral modes of the vehicle. These newly generated behavior nodes, which may exhibit significant interdependencies, are interconnected through edges $e_{ij}^b$, forming a more complex interaction structure. Consequently, the adjacency matrix $A^b$ is extended to accommodate the expanded node set, resulting in increased dimensionality. Here, $\Lambda$ is a behavior correlation matrix, and each element $\Lambda_{mn}$ in the matrix represents the correlation between behavior mode $m$ of one agent and behavior mode $n$ of another agent. The feature tensor $X^b$ of the behavior nodes encodes the possible motion states under each behavioral mode, capturing the multi-modal nature of vehicle behavior.

\begin{figure}[!ht]
\centering
\includegraphics[width=0.85\textwidth]{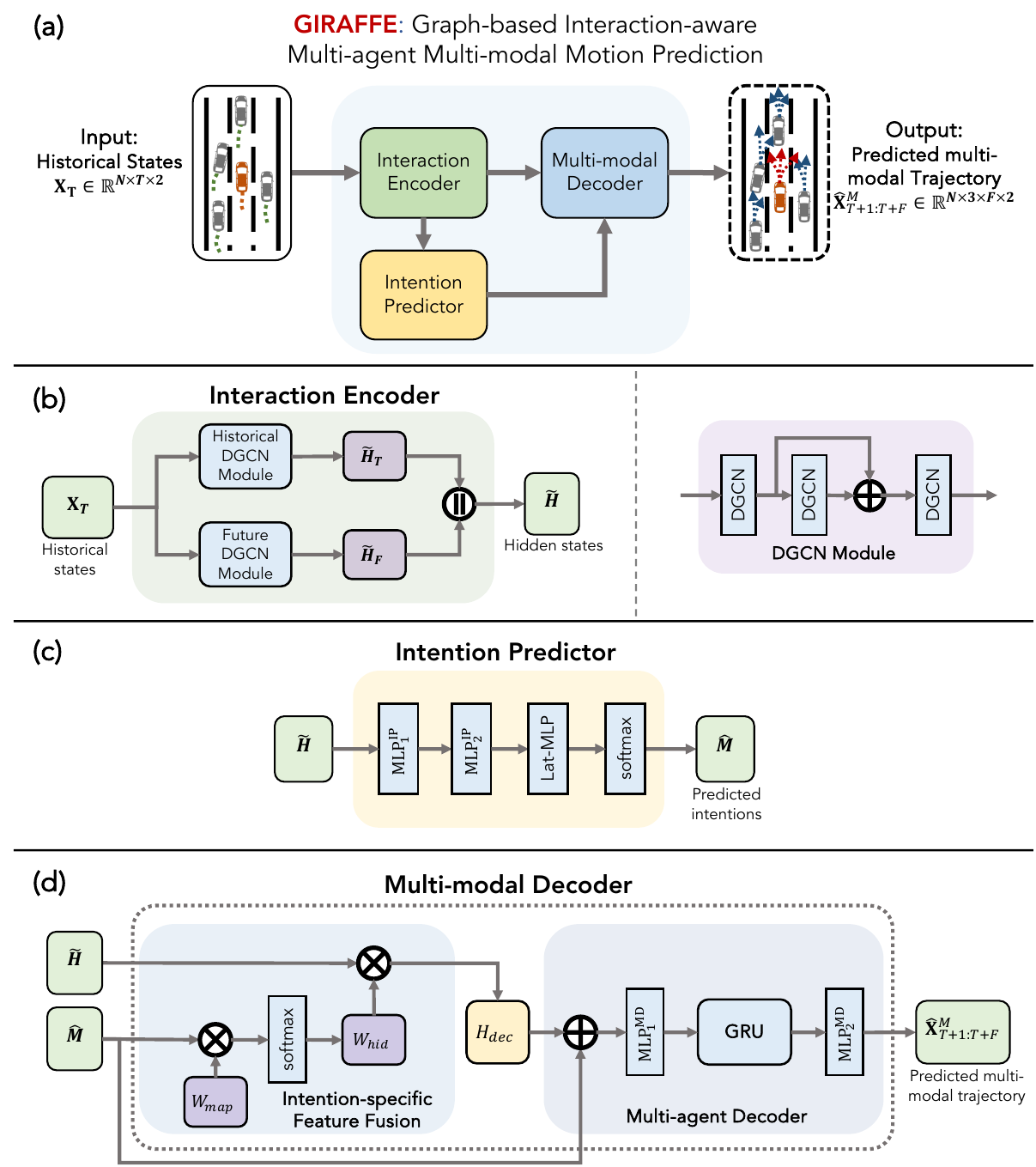}
\caption{GIRAFFE Framework.}
\label{fig_GIRAFFE}
\end{figure}

Based on this foundation, we instantiate a deep neural architecture following \cite{wu2023graph} that (i) encodes interactions between the target vehicle and its surroundings via the agent graph $\mathcal{G}^a$, and (ii) expresses its multi-agent, multi-modal trajectory hypotheses as an agent–behavior graph $\mathcal{G}^b$, as illustrated in Figure \ref{fig_GIRAFFE}. The predictor $\mathbf{H}^{Pred}(\cdot)$ comprises three core modules, detailed below.

\paragraph{\textbf{Interaction Encoder}} 
The Interaction Encoder utilizes a Diffusion Graph Convolution Network (DGCN) architecture to encode dynamic spatial relations among vehicles, as described in \ref{app_dgcn}.  Unlike a standard GCN, a DGCN performs separate ``forward'' and ``reverse'' diffusion convolutions over the adjacency matrix, explicitly modeling how influences propagate downstream and upstream \cite{li2017diffusion,wu2021inductive}.  By leveraging Chebyshev polynomial filters \(T_k(\cdot)\) on both the normalized forward \(\bar A^f\) and backward \(\bar A^b\) transition matrices, DGCN can aggregate information from \(k\)-hop neighborhoods in a bidirectional manner without requiring deep stacks of layers.  This choice ensures that our encoder captures asymmetric, higher-order interactions essential for modeling realistic highway dynamics. We employ two parallel DGCN branches. The first branch, \(DGCN_H\), encodes the historical observations \(X_T\) into the past embedding \(\tilde H_T\).  The second branch, \(DGCN_F\), takes the same historical inputs but is tasked with predicting the vehicles’ future 2D feature sequences. Formally, we concat the two outputs to derive a joint embedding $\tilde H$:

\begin{equation}
    \tilde H_T = \mathrm{DGCN_H}(X_T)
\end{equation}
\begin{equation}
    \tilde H_F = \mathrm{DGCN_F}(X_T)
\end{equation}
\begin{equation}
    \tilde H = [\tilde H_T, \tilde H_F]
\end{equation}

\paragraph{\textbf{Intention Predictor}} The Intention Predictor addresses the classification of future driving intentions, both laterally and longitudinally. Using the encoded graph representation, $\mathcal{F}_i$ which contains two MLP layers reduce the dimensions and encode the features into a latent space. The MLP layer $\mathcal{F}_{lat}$ with softmax activation then classify the three lateral intentions over the future time horizon. By conditioning this classifier on DGCN-derived spatial features, the network learns context-aware intentions and helps in understanding the potential maneuvers the vehicle might take, such as lane changes or speed adjustments.

\begin{equation}
    \tilde H^{IP} = \mathcal{F}_i (\tilde H)
\end{equation}
\begin{equation}
    \hat M
    = \mathrm{softmax}\Bigl(\mathcal{F}_{lat}\bigl(\tilde H^{IP}\bigr)\Bigr)
\end{equation}

\paragraph{\textbf{Multi-modal Decoder}} Finally, the Multi-modal Decoder fuses the predicted intentions of multiple agents with the latent space to produce multiple future trajectory distributions for each agent. This decoder uses a trainable weight matrix to combine features from distinct historical and future time steps, emphasizing the importance of sequential motion patterns. The GRU-based decoder $\mathcal{F}_d$ with two MLP layers ensures temporal continuity in the predicted trajectories, mapping the fused features to a bivariate Gaussian distribution representing the future vehicle positions. This approach allows the model to generate probabilistic predictions for multiple agents.

\begin{equation}
    W_{hid} = \mathrm{softmax}( W_{map} \otimes \hat M)
\end{equation}
\begin{equation}
    H_{dec} = W_{hid} \cdot \tilde H
\end{equation}
\begin{equation}
    \hat{\mathbf{X}}_{T+1:T+F}^M  = \mathcal{F}_d (H_{dec}, \hat M)))
\end{equation}

\paragraph{\textbf{Construction of Agent–Behavior Graph}}
After the Multi‐modal Decoder produces the set of \(M\) trajectory hypotheses 
\(\hat{\mathbf{X}}_{T+1:T+F}^M\), we assemble \(\mathcal{G}^b\) dynamically.  
Each agent node \(v_i^a\) is cloned into behavior nodes \(v^b_{i,m}\) for \(m\in\{L,K,R\}\), with features
\begin{equation}
  X_{i,m}^b
  = \bigl[\;\tilde H_i;\;\hat m_{i,m};\;\hat{\mathbf{x}}_{T+1:T+F}^{m,i}\bigr],
\end{equation}
where \(\hat{\mathbf{x}}_{T+1:T+F}^{m,i}\) is the \(i\)-th agent’s \(m\)-th mode trajectory from \(\hat{\mathbf{X}}_{T+1:T+F}^M\).  
The adjacency is then lifted via
\begin{equation}
  A^b = A^a \otimes \Lambda,
\end{equation}
with \(\Lambda\in\mathbb{R}^{3\times3}\) encoding mode correlations.  This graph captures both the multi‐modal outputs and their inter‐agent context for downstream supervision and reasoning.

\subsection{RHINO: Hypergraph-based Motion Generator}


Unlike traditional graph representations, which are confined to pair-wise relationships, hypergraphs offer a more sophisticated and comprehensive framework for representing group-wise interactions. By connecting multiple vehicles that exhibit strong correlations through hyperedges, hypergraphs enable a more robust analysis and optimization of the complex network of interactions.  The concept of an Agent Hypergraph to represent agents and their group-wise interactions is introduced as follows:

\begin{definition} [Agent Hypergraph] \label{agent_hypergraph}
    Let $\mathcal{H}^b$ be a hypergraph representation of the motion states of $N$ agents, with each agent represented as a node. The hypergraph $\mathcal{H}^a$ is expressed as 
    $$\mathcal{H}^a=(\mathcal{V}^a, \mathcal{U}^a; X^a, H^a
    )$$
    where $\mathcal{V}^a \in \mathbb{R}^N$ denotes the node set, $\mathcal{U}^a \in \mathbb{R}^L$ denotes the edge set, $X^a \in \mathbb{R}^{N \times C}$ represents the feature tensor, $H^a \in \mathbb{R}^{N \times L}$ indicates the incidence matrix, where $H^a_{ij}$ indicates whether node $v_i$ is part of the hyperedge $u_j$.
\end{definition}

\begin{figure}[!ht]
\centering
\includegraphics[width=0.9\textwidth]{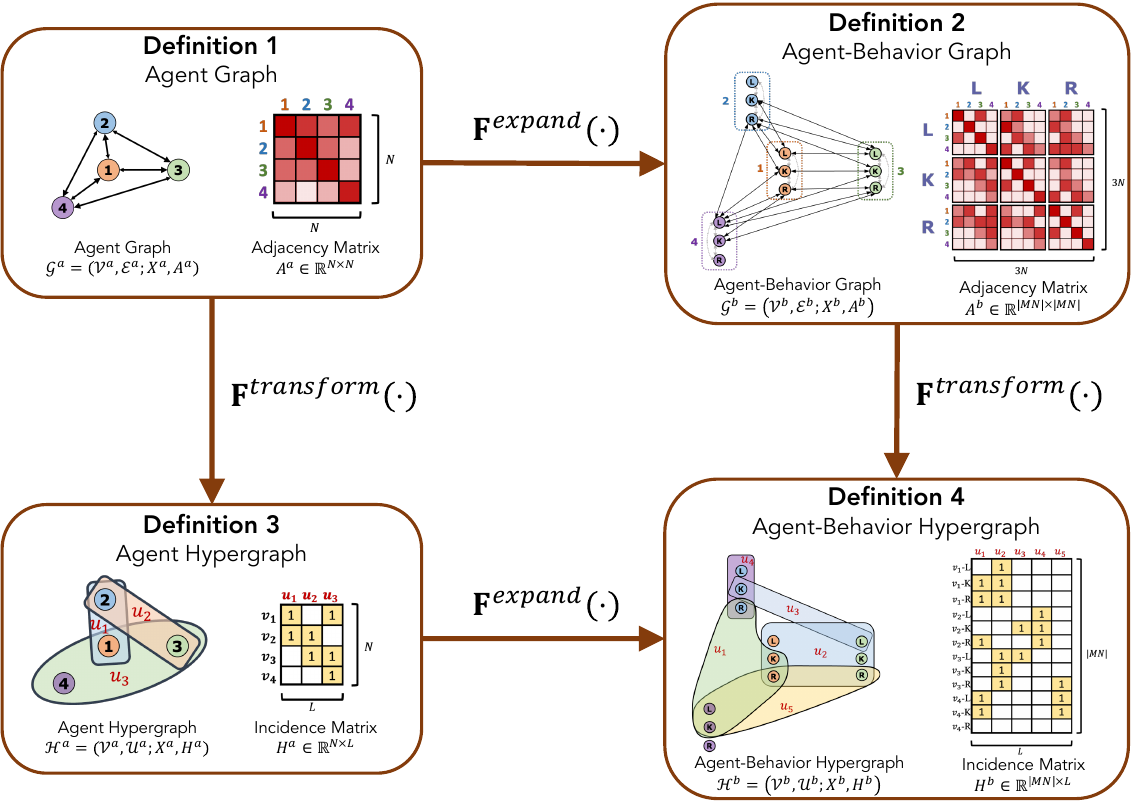}
\caption{Definitions of graphs and hypergraphs}
\label{fig_hypergraph_def}
\end{figure}

To convert an Agent Graph $\mathcal{G}^a$ into an Agent Hypergraph $\mathcal{H}^a$, we introduce a transformation function $\mathbf{F}^{transform}(\cdot)$. This function enables the shift from a pairwise interaction framework to a higher-order interaction model represented by the hypergraph. Formally, the transformation is expressed as:

\begin{equation} 
    \mathcal{H}^{a} = \mathbf{F}^{transform} \left(\mathcal{G}^a \right)
    \Leftrightarrow 
    \left\{ 
    \begin{aligned} 
        \mathcal{V}^a &= \bigcup_{i=1}^{N} \{ v_i^a \} \\ 
        \mathcal{U}^a &= \bigcup_{j=1}^{K} \left\{ u_j^a \mid u_j^a \subseteq \mathcal{V}^a, \Lambda_{u_j} \neq 0 \right\} \\
        X^a &= \bigcup_{i=1}^{N} X_{i}^a \\ 
        H^a &= \bigcup_{i=1}^{N} \bigcup_{j=1}^{K} \left\{ H^a_{ij} \mid v_i^a \in u_j^a \right\}
    \end{aligned} 
    \right.
\end{equation}

In this transformation, the node set $\mathcal{V}^a$ and the feature tensor $X^a$ remain consistent between the graph and the hypergraph representations. However, the primary modification occurs in the edge formulation. The transformation replaces the pairwise edges of the original Agent Graph with hyperedges that can connect multiple nodes simultaneously. This redefinition of edges as hyperedges within the hypergraph $\mathcal{H}^a$ allows for the modeling of group-wise interactions, where a single hyperedge $u \in \mathcal{U}^a$ can link more than two nodes, capturing higher-order relationships among agents. The incidence matrix $H^a$ is updated to reflect this change, where each entry $H^a_{ij}$ indicates whether agent $v_i$ participates in hyperedge $u_j$. As a result, $\mathcal{H}^a$ can represent multi-agent interactions that involve multiple agents simultaneously, providing a richer and more flexible structure for modeling the dynamics of the system.

To enhance the understanding of complex group-wise interactions in multi-agent systems, it is essential to extend the traditional Agent Hypergraph model to account for the diverse behavioral modes of each agent. This is achieved by decomposing each agent node in the Agent Hypergraph $\mathcal{H}^a$ into multiple behavior-specific nodes, which correspond to the different modes of behavior each agent can exhibit. The result of this decomposition is the Agent-Behavior Hypergraph, denoted as $\mathcal{H}^{b}$.

\begin{definition} [Agent-Behavior Hypergraph] \label{agent_behavior_hypergraph}
    Let $\mathcal{H}^b$ be a hypergraph representation of the multi-modal motion states of $N$ agents, with each of $M$ behavior modes for each agent represented as a node. The hypergraph $\mathcal{H}^b$ is expressed as 
    $$\mathcal{H}^b=(\mathcal{V}^b, \mathcal{U}^b; X^b, H^b)$$
    where $\mathcal{V}^b \in \mathbb{R}^{|MN|}$ denotes the node set, $\mathcal{U}^b \in \mathbb{R}^L$ denotes the edge set, $X^b \in \mathbb{R}^{|MN| \times C}$ represents the feature tensor, $H^b \in \mathbb{R}^{|MN| \times L}$ indicates the incidence matrix, where $H_{ij}^b$ indicates whether node $v_i$ is part of the hyperedge $u_j$.
\end{definition}

To formally describe the process of transitioning from an Agent Hypergraph $\mathcal{H}^{a}$ to an Agent-Behavior Hypergraph $\mathcal{H}^{b}$, the expansion function $\mathbf{F}^{expand}(\cdot)$ is applied. This function decomposes each agent node into multiple behavior-specific nodes and updates the hyperedge structure accordingly. The behavior-specific nodes correspond to the different behavior modes, while the hyperedges represent the higher-order interactions among the behavior modes of different agents.

\begin{equation} 
    \mathcal{H}^{b}  = \mathbf{F}^{expand} \left( \mathcal{H}^a \right)
    \Leftrightarrow 
    \left\{ 
    \begin{aligned} 
        \mathcal{V}^b 
        &= \bigcup_{i=1}^{N} \bigcup_{m=1}^{M} \left\{ v_{i,m}^b \right\} \\ 
        \mathcal{U}^b &= \bigcup_{j=1}^{L} \bigcup_{m=1}^{M} \bigcup_{n=1}^{M} \left\{ u_{j,mn}^b \mid u_j^a \neq 0 \text{ and } \Lambda_{mn} \neq 0 \right\} \\ 
        X^b &= \bigcup_{i=1}^{N} \bigcup_{m=1}^{M} X_{i,m}^a \\ 
        H^b &= \bigcup_{i=1}^{MN} \bigcup_{j=1}^{L} \left\{ H^b_{ij} \mid v_{i,m}^b \in u_{j,mn}^b \right\} 
    \end{aligned} 
    \right. 
\end{equation}

The node set $\mathcal{V}^b$ expands each agent $v_{i}^b$ into multiple behavior-specific nodes $v_{i,m}^b$, where each $m$ represents a different behavioral mode of the agent. The hyperedges $\mathcal{U}^b$ are formed between the behavior-specific nodes based on the group-wise interactions present in the original hypergraph, with the correlation between behavior modes captured by $\Lambda_{mn}$. The feature tensor $X^b$ captures the state of each behavior node, inheriting the feature data from the original hypergraph.The incidence matrix $H^b$ records whether a behavior-specific node $v_{i,m}^b$ is part of a hyperedge $u_{j,mn}^b$. In this extended framework, hyperedges can represent these aforementioned complex group-wise interactions by connecting behaviors of multiple vehicles that are influenced simultaneously by a shared context, as illustrated in Figure \ref{fig_hypergraph_def}. Therefore, an Agent-Behavior Hypergraph is defined to model the multi-agent, multi-modal system for reasoning about group-wise interaction relations.

In addition to the expansion process, the transformation function $\mathbf{F}^{transform}(\cdot)$ converts an Agent-Behavior Graph $\mathcal{G}^b$ into an Agent-Behavior Hypergraph $\mathcal{H}^b$. This transformation replaces the pairwise edges of the graph with hyperedges that capture higher-order interactions between behavior modes across multiple agents. The transformation is guided by the adjacency matrix $A^b$ of the original graph and the behavior-mode correlation matrix $\Lambda_{mn}$. The transformation function $\mathbf{F}^{transform}(\cdot)$ is expressed as:
\begin{equation} 
    \mathcal{H}^{b} = \mathbf{F}^{transform} \left(
    \mathcal{G}^b \right)
    \Leftrightarrow \left\{ 
    \begin{aligned} 
        \mathcal{V}^b 
        &= \bigcup_{i=1}^{N} \bigcup_{m=1}^{M} \left\{ v_{i,m}^b \right\} \\ 
        \mathcal{U}^b &= \bigcup_{j=1}^{L} \bigcup_{m=1}^{M} \bigcup_{n=1}^{M} \left\{ u_{j,mn}^b \mid A^b_{ij} \neq 0 \text{ and } \Lambda_{mn} \neq 0 \right\} \\ 
        X^b &= \bigcup_{i=1}^{N} X_{i}^b \\ 
        H^b &= \bigcup_{i=1}^{MN} \bigcup_{j=1}^{L} \left\{ H^b_{ij} \mid v_{i,m}^b \in u_{j,mn}^b \right\} 
    \end{aligned} 
    \right. 
\end{equation}

This structure allows for the connection of behavior nodes across different agents, enabling the representation of interactions among diverse behaviors of multiple agents in a shared context. Hence, the Agent-Behavior Hypergraph not only captures the individual behavior of each agent but also models how these behaviors interact and influence one another within a multi-agent system.

\subsubsection{\texttt{RHINO} Framework Architecture}
The core of \texttt{RHINO} is to learn a multi-scale Agent-Behavior Hypergraph, where nodes represent the behaviors of agents and hyperedges capture their group-wise interactions. This hypergraph is then used to learn agent and interaction embeddings to better understand the underlying interaction relations. We also incorporate a basic multi-agent trajectory generation system based on the CVAE framework to handle the stochasticity of each agent's potential behaviors and motion states, generating plausible trajectories for each vehicle.

\begin{figure}[!ht]
\centering
\includegraphics[width=0.96\textwidth]{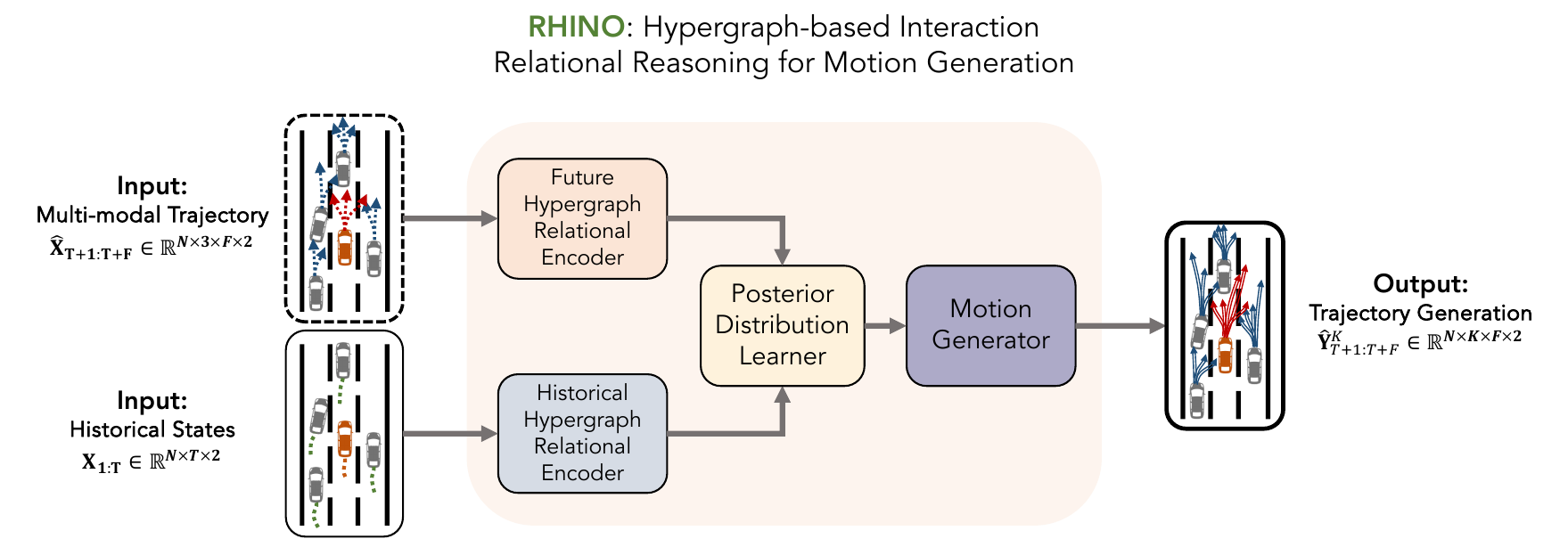}
\caption{RHINO Framework.}
\label{fig_RHINO}
\end{figure}

Thus, as illustrated in Figure \ref{fig_RHINO}, \texttt{RHINO} comprises the following modules:
\begin{itemize}
    \item \textbf{Hypergraph Relational Encoder}, which transforms both the original historical states and predicted multi-agent multi-modal trajectories into hypergraphs, modeling and reasoning the underlying relation between the vehicles.
    \item \textbf{Posterior Distribution Learner}, which captures the posterior distribution of the future trajectory given the historical states and the predicted multi-modal future motion states of all the vehicles in the vehicle group.
    \item \textbf{Motion Generator}, which decodes the embeddings by concurrently reconstructing the historical states and generating the future trajectories.
\end{itemize}

\subsubsection{Hypergraph Relational Encoder}

We employ two Hypergraph Relational Encoder modules: a Historical Hypergraph Relational Encoder for handling historical states and a Future Hypergraph Relational Encoder for predicted multi-agent multi-modal trajectories from \texttt{GIRAFFE}. For the Historical Hypergraph Relational Encoder, the input historical states $X_T$ form an Agent Hypergraph $\mathcal{H}_T^a$. For the Future Hypergraph Relational Encoder, the predicted multi-agent multi-modal trajectories $\hat X_{T+1:T+F}$ form an Agent-Behavior Hypergraph $\mathcal{H}_F^b$, where each agent node is expanded into three lateral behavior nodes with corresponding predicted future states. Both modules share the same structure regardless of the input hypergraph types.

\begin{figure}[!ht]
\centering
\includegraphics[width=0.85\textwidth]{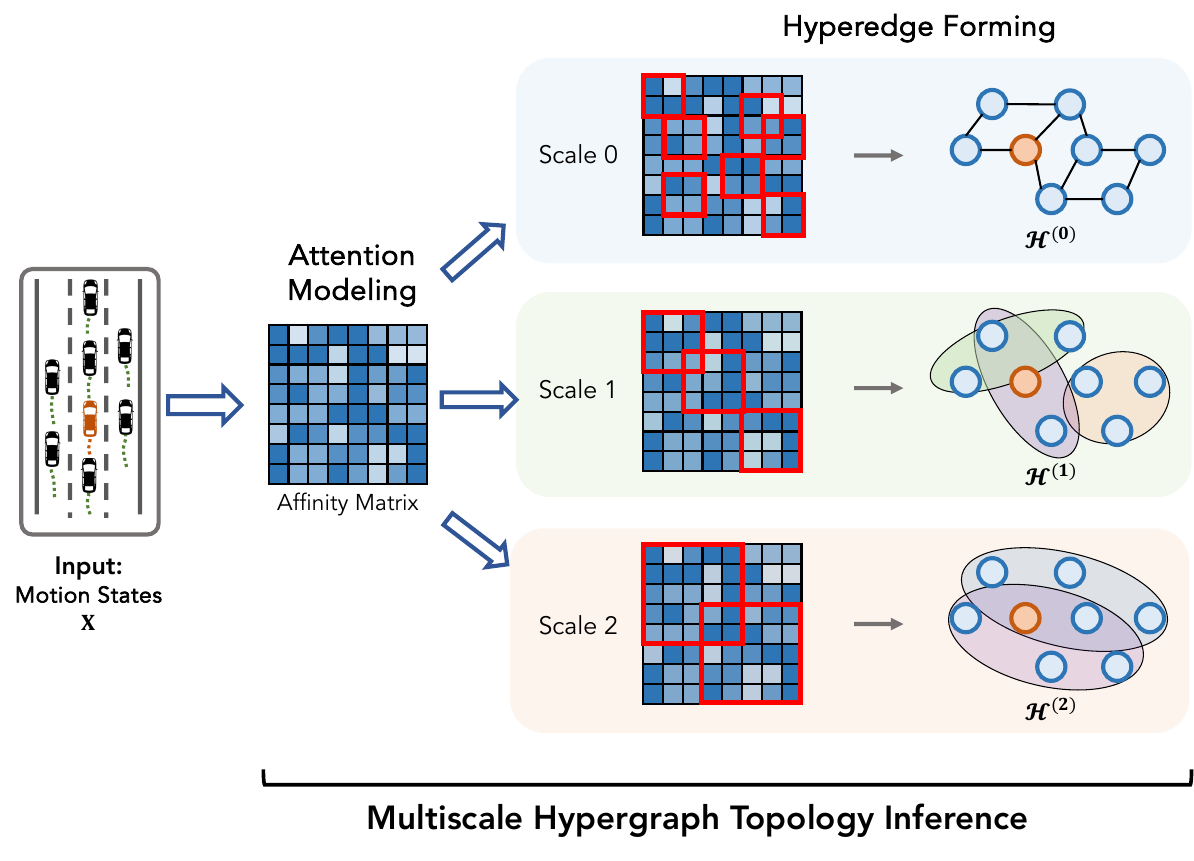}
\caption{Multi-scale Hypergraph Topology Inference.}
\label{fig_hypergraph_encoder_1}
\end{figure}

\paragraph{\textbf{Multi-scale Hypergraph Topology Inference}}
Real‐world traffic interactions occur at multiple levels of granularity, so a single hyperedge size cannot capture the full spectrum of group dynamics. To comprehensively model group-wise interactions in the hypergraphs at multiple scales, we infer a hierarchical family of hypergraphs \(\{\mathcal{H}^{(s)}\}_{s=0}^S\). Here, \(\mathcal{H}^{(0)}\) models the finest pairwise connections, and as the scale \(s\) increases, each hyperedge grows to size \(\lvert u\rvert = J^{(s)}\), capturing progressively larger vehicle groups. As shown in Figure \ref{fig_hypergraph_encoder_1}, at any scale $s$, $\mathcal{H}^{(s)} = (\mathcal{V}, \mathcal{U}^{(s)})$ has a node set $\mathcal{V} = \{v_1, v_2, \cdots, v_N\}$ and a hyperedge set $\mathcal{U}^{(s)} = \{u_1^{(s)}, u_2^{(s)}, \cdots, u_{J}^{(s)}\}$ representing group-wise relations with $J$ hyperedges. The topology of each $\mathcal{H}^{(s)}$ is represented as an incidence matrix $H^{(s)}$.

To understand and quantify the dynamic interactions between agents within a given system, we adopt trajectory embedding to distill the motion states of agents into a compact and informative representation. To infer a multi-scale hypergraph, we construct hyperedges by grouping agents that have highly correlated trajectories, whose correlations could be measured by mapping the trajectories as a high-dimensional feature vector. For the $i$-th agent in the system, the trajectory embedding is denoted as $q_i$. This embedding is a function of the agent’s motion states, defined over a temporal window extending from time 1 to time $T$. The embedding function $f_q$, which is a trainable MLP, is responsible for transforming the motion states $X^i$ into a vector $q_i \in \mathbb{R}^d$, where $d$ is the dimensionality of the embedded space. Mathematically, the trajectory embedding is represented as:
\begin{equation}
    q_i = f_q (X^i)
\end{equation}

The affinity between agents is represented by an affinity matrix $\mathcal{A} \in \mathbb{R}^{N \times N}$, which contains the pairwise relational weights between all agents. The affinity matrix is defined as:

\begin{equation}
    \mathcal{A} = \{ \mathcal{A}_{ij} | \forall i, j = 1, \dots, N \}
\end{equation}

Each element $\mathcal{A}_{ij}$ is computed as the correlation between the trajectory embeddings of the $i$-th and $j$-th agents. The correlation is the normalized dot product of the two trajectory embeddings, expressed as:

\begin{equation}
    \mathcal{A}_{ij} = \frac{q_i^\top q_j}{\|q_i\|_2 \|q_j\|_2}
\end{equation}

Here, $\|\cdot\|_2$ denotes the L2 norm. The relational weight $\mathcal{A}_{ij}$ measures the strength of association between the trajectories of the $i$-th and $j$-th agents, capturing the degree to which their behaviors are correlated. This enables the assessment of interaction patterns and can uncover underlying social or physical laws governing agent dynamics.

The formulation of a hypergraph necessitates the strategic formation of hyperedges that reflect the complex interaction between the nodes in the system. At the outset, the 0-th scale hypergraph $\mathcal{H}^{(0)}$ is considered, where the construction is based on pair-wise connections. Each node establishes a link with another node that has the highest affinity score with it. 

As the complexity of the system is scaled up, beginning at scale $s \geq 1$, the methodology shifts towards group-wise connections. This shift is based on the intuition that agents within a particular group should display strong mutual correlations, suggesting a propensity for concerted action. To implement this, a sequence of increasing group sizes $\{ J^{(s)} \}_{s=1}^S$ is established. For every node, denoted by $v_i$, the objective is to discern a group of agents that are highly correlated, ultimately leading to $J^{(s)}$ groups or hyperedges at each scale $s$. The hyperedge associated with a node $v_i$ at a given scale $s$ is indicated by $u_i^{(s)}$. The determination of the most correlated agents is framed as an optimization problem, aiming to link these agents into a hyperedge that accounts for group dynamics:

\begin{equation}
    u_i^{(s)} = \underset{\Gamma \subseteq \mathcal{V}}{\mathrm{arg max}} \left\| \mathcal{A}_{\Gamma,\Gamma} \right\|_{1} 
\end{equation}
\begin{equation}
    \text{s.t.} \; |\Gamma| = J^{(s)}; v_i \in \Gamma; i=1, \ldots, N
\end{equation}

The culmination of this hierarchical structuring is a multi-scale hypergraph, encapsulated by the set $\{ \mathcal{H}^{(s)} \in \mathbb{R}^{N \times N} \}_{s=1}^S$, where each scale  $s$ embodies a distinct layer of abstraction in the representation of node relationships within the hypergraph.

\begin{figure}[!ht]
\centering
\includegraphics[width=0.9\textwidth]{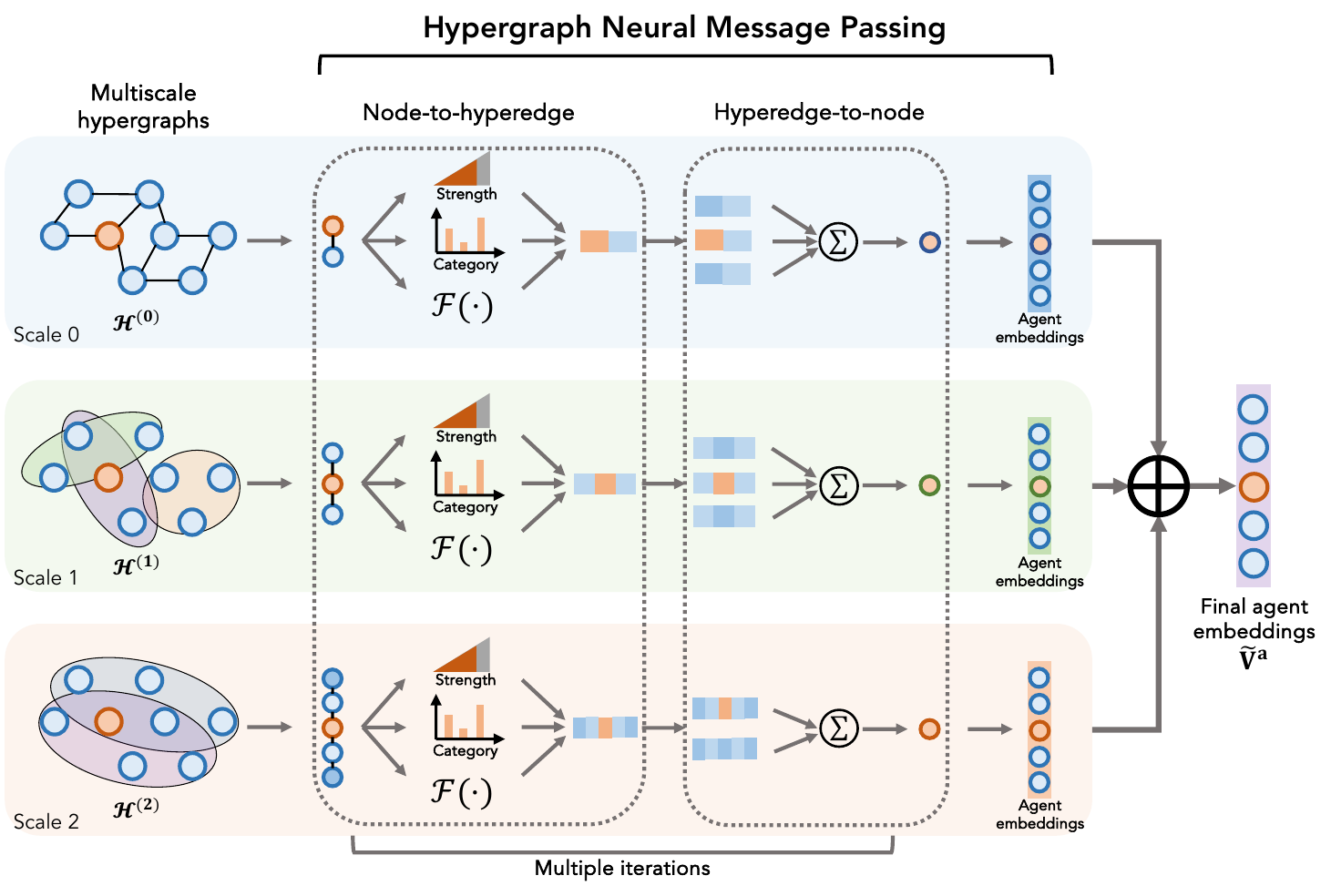}
\caption{Hypergraph Neural Message Passing.}
\label{fig_hypergraph_encoder_2}
\end{figure}

\paragraph{\textbf{Hypergraph Neural Message Passing}}
In order to discern the patterns of agent motion states from the inferred multi-scale hypergraph, we have tailored a multi-scale hypergraph neural message passing technique to construct the hypergraph topology. This method iteratively acquires the embeddings of vehicles and the corresponding interactions through node-to-hyperedge and hyperedge-to-node processes, as depicted in Figure \ref{fig_hypergraph_encoder_2}.  

The node-to-hyperedge mapping aggregates agent embeddings to generate interaction embeddings. Initially, for any given scale, the initial embedding for the $i$th agent, $v_i = q_i \in \mathbb{R}^d$. Each node $v_j$ is associated with a hyperedge $u_i$, given that $v_j$ is an element of $u_i$. This mapping facilitates the definition of the hyperedge interaction embedding. The hyperedge interaction embedding for a hyperedge $u_i$ is defined as a function of the embeddings of the nodes contained within it, modulated by the neural interaction strength $r_i$ and categorized through coefficients $c_{i,l}$. The per-category function $\mathcal{F}_l$ models the interaction process for each category, which is crucial for capturing the nuances of different interaction types. Each $\mathcal{F}_l$ is a trainable MLP, responsible for processing the aggregated node embeddings within the context of a specific interaction category. The mathematical formulation is:

\begin{equation}
    u_i = r_i \sum_{l=1}^L c_{i,l} \mathcal{F}_l \left( \sum_{v_j \in u_i} v_j \right)
\end{equation}

The neural interaction strength $r_i$ encapsulates the intensity of the interaction within the hyperedge and is obtained through a trainable model $\mathcal{F}_r$, applied to a collective embedding $z_i$ with a sigmoid function $\sigma$ as Eq.(\ref{eq21}). This collective embedding $z_i$ is represented as the weighted sum of the individual node embeddings within the hyperedge, signifying the aggregated information of agents in a group as Eq. (\ref{eq22}). The weight $w_j$ for each node is determined by a trainable MLP $\mathcal{F}_w$ as Eq.\ref{eq23}.

\begin{equation} \label{eq21}
    r_i = \sigma(\mathcal{F}_r (z_i))
\end{equation}
\begin{equation} \label{eq22}
    z_i = \sum_{v_j \in u_i} w_j v_j
\end{equation}
\begin{equation} \label{eq23}
    w_j = \mathcal{F}_w \left( v_j, \sum_{v_m \in u_i} v_m \right)
\end{equation}

The neural interaction category coefficient $c_i$ represent the model's reasoning about which type of the interaction is likely for hyperedge $u_j$, where $c_{i,l}$ denotes the probability of the $l$-th neural interaction category within $L$ possible categories. These coefficients are computed using a softmax function applied to the output of another trainable MLP $\mathcal{F}_c$, which is further adjusted by an i.i.d. sample Gumbel distribution $g$ as described in \ref{app_gumbel}.  which add some random noise and a temperature parameter $\tau$ which controlls the smoothness of probability distribution \cite{jang2016categorical}:

\begin{equation}
    c_i = \mathrm{softmax}\left(\frac{\mathcal{F}_c (z_i) + g}{\tau}\right)
\end{equation}

These components, including neural interaction strength, interaction category coefficients, and per-category functions, provide a comprehensive mechanism for reasoning over complex, higher-order relationships, allowing the model to adapt its understanding of how agents collectively behave in diverse scenarios.

The process of hyperedge-to-node mapping is a pivotal step that allows for the update and refinement of agent embeddings within the hypergraph framework. Each hyperedge $u_j$ is mapped back onto its constituent nodes $v_i$, assuming every $v_i$ is included in $u_j$. The primary objective of this phase is to update the embedding of an agent. This is achieved through the function $\mathcal{F}_v$, which is a trainable MLP. The updated agent embedding $\tilde v_i$ is the result of the function applied to the concatenation of the agent's current embedding and the sum of the embeddings of all hyperedges that the agent is part of. Formally, the update rule for the agent embedding is represented as:

\begin{equation}
    \tilde v_i \leftarrow \mathcal{F}_v \left( \left[ \, v_i, \sum_{u_j \in \mathcal{U}_i} u_j \, \right]\right)
\end{equation}
where $\mathcal{U}_i = \{u_j \,|\, v_i \in u_j\}$ denotes the set of hyperedges associated with the $i$-th node $v_i$, and $[\, \cdot, \cdot \,]$ symbolize the operation of embedding concatenation. This operation fuses the individual node embedding with the collective information conveyed by the associated hyperedges. This amalgamation is crucial as it encapsulates the influence exerted by the interactions within the hyperedges onto the individual agent.

The Hypergraph Relational Encoder applies the node-to-hyperedge and hyperedge-to-node phases iteratively, allowing agent embeddings to be refined and enriched as relationships within hyperedges evolve. Upon the completion of these iterations, the output is constructed as the concatenation of the agent embeddings across all scales. The final agent embedding matrix $\mathbf{\tilde V}^a$ is composed of the embeddings of all agents, where each agent embedding $v_i$ is a concatenation of the embeddings from all scales, expressed as:

\begin{equation}
    \mathbf{\tilde V}^a = [ \tilde  v_i ] \in \mathbb{R}^{N \times |d(S+1)|}, \quad \forall i \in [1, \ldots, N] 
\end{equation}
where
\begin{equation}
    \tilde v_i = [ \tilde v_i^{(0)}, \tilde v_i^{(1)}, \ldots, \tilde v_i^{(S)} ]
\end{equation}

By stacking message‐passing over these scales, the final node embedding integrates both pair-wise and group-wise interactions, enabling \texttt{RHINO} to reason about tight and large‐scale vehicle groups in one unified framework.

\subsubsection{Posterior Distribution Learner}
\begin{figure}[!ht]
\centering
\includegraphics[width=0.95\textwidth]{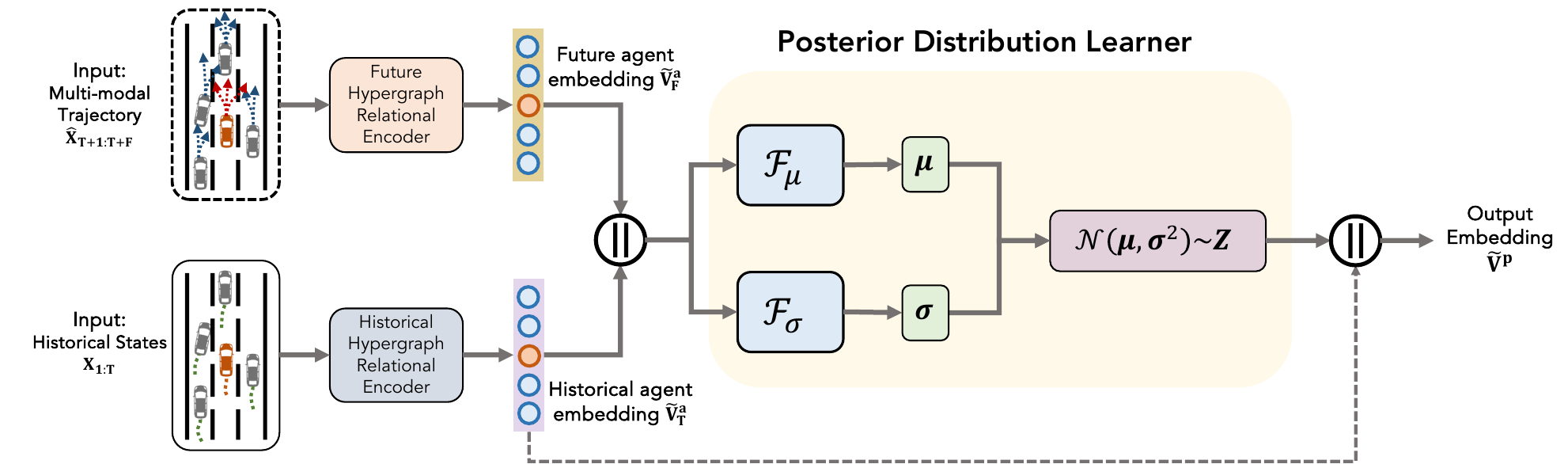}
\caption{Posterior Distribution Learner.}
\label{fig_posterior_distribution_learner}
\end{figure}
In our study, we incorporated multi-scale hypergraph embeddings into a multi-agent trajectory generation system using the CVAE framework \cite{pagnoni2018conditional} to address the stochastic nature of each agent’s behavior, as shown in Figure \ref{fig_posterior_distribution_learner}. Here, we denote the historical trajectories $\mathbf{X}_{1:T}$ as $\mathbf{X}_T$, and denote the predicted future trajectories $\mathbf{X}_{T+1:T+F}$ as $\mathbf{X}_F$. Let $\log p(\mathbf{X}_F | \mathbf{X}_T)$ denote the log-likelihood of predicted future trajectories $\mathbf{X}_F$ given historical trajectories $\mathbf{X}_T$. The core challenge is that, even with identical histories, multiple future trajectories are plausible due to unobserved driver intentions, sensor noise, and varying traffic conditions.  A CVAE provides a principled way to learn a continuous latent space \(\mathbf{Z}\in\mathbb{R}^{N\times d_z}\) that captures this multi‐modality and quantifies predictive uncertainty. The corresponding Evidence Lower Bound (ELBO) is defined as follows:
\begin{equation}
    \begin{split}
        \log p(X_F | X_T) 
        &\geq \mathbb{E}_{q_\phi(\mathbf{Z} | \mathbf{X}_F,\mathbf{X}_T)} \log p_\theta(\mathbf{X}_F | \mathbf{Z}, \mathbf{X}_T) \\
        &- \text{KL}(q_\phi(\mathbf{Z} | \mathbf{X}_F, \mathbf{X}_T) \parallel p_\psi(\mathbf{Z} |\mathbf{X}_T)),
    \end{split}
\end{equation}
where $\mathbf{Z} \in \mathbb{R}^{N \times d_z}$ represents the latent codes corresponding to all agents; $p_\psi(\mathbf{Z} | \mathbf{X}_T)$ is the conditional prior of $\mathbf{Z}$, modeled as a Gaussian distribution. $\text{KL}$ represents the Kullback–Leibler divergence function. In this framework, $q_\phi(\mathbf{Z} | \mathbf{X}_F, \mathbf{X}_T)$ is implemented through an encoding process for embedding learning, and $p_\theta(\mathbf{X}_F | \mathbf{Z}, \mathbf{X}_T)$ is realized via a decoding process that forecasts the future trajectories $\mathbf{X}_F$. In this loss, the first term (reconstruction) encourages the decoder \(p_\theta\) to produce futures \(\hat{\mathbf{X}}_F\) close to ground truth, while the KL term regularizes the posterior \(q_\phi\) toward the prior \(p_\psi\), preventing over‐confident, narrow distributions and thus embedding calibrated uncertainty in \(\mathbf{Z}\).

Thus, the goal of the Posterior Distribution Learner is to derive the Gaussian parameters for the approximate posterior distribution. This involves computing the mean $\mu_q$ and the variance $\sigma_q$ based on the final output embeddings $\mathbf{\tilde V}_F^a$ and the target embeddings $\mathbf{\tilde V}_T^a$. These parameters are generated through two separate trainable MLPs, $\mathcal{F}_\mu$ and $\mathcal{F}_\sigma$, respectively. The latent code $Z$, representing possible trajectories, is then sampled from a Gaussian distribution parameterized by these means and variances. The final output embeddings $\mathbf{\tilde V}^{p}$ are a concatenation of the latent code $\mathbf{Z}$, the final output embeddings $\mathbf{\tilde V}_F^a$, and the target embeddings $\mathbf{\tilde V}_T^a$. The equations governing these processes are as follows:

\begin{equation}
    \mu_q = \mathcal{F}_\mu (\mathbf{\tilde V}_F^a, \mathbf{\tilde V}_T^a)
\end{equation}
\begin{equation}
    \sigma_q = \mathcal{F}_\sigma (\mathbf{\tilde V}_F^a, \mathbf{\tilde V}_T^a)
\end{equation}
\begin{equation}
    \mathbf{Z} \sim \mathcal{N}(\mu_q, \text{Diag}(\sigma_q^2))
\end{equation}
\begin{equation}
    \mathbf{\tilde V}^{p} = [\mathbf{Z}, \mathbf{\tilde V}_T^a]
\end{equation}

In these notations, $\mu_q$ and $\sigma_q$ represent the mean and variance of the approximated posterior distribution. $\mathcal{F}_\mu$ and $\mathcal{F}_\sigma$ are the trainable MLPs that produce these parameters. $\mathbf{Z}$ denotes the latent code of possible trajectories, and $\mathbf{\tilde V}^{p}$ stands for the output embeddings, which fuses the latent code and the historical embeddings. At inference, sampling from the learned prior \(p_\psi(\mathbf{Z}\mid\widetilde V_T^a)\) and decoding yields \(K\) diverse trajectory hypotheses, each reflecting different plausible futures under the same observed history.  By embedding uncertainty directly in \(\mathbf{Z}\), \texttt{RHINO}’s Motion Generator can robustly capture and propagate the inherent variability in multi‐agent driving scenarios.

\subsubsection{Motion Generator}

\begin{figure}[!ht]
\centering
\includegraphics[width=0.8\textwidth]{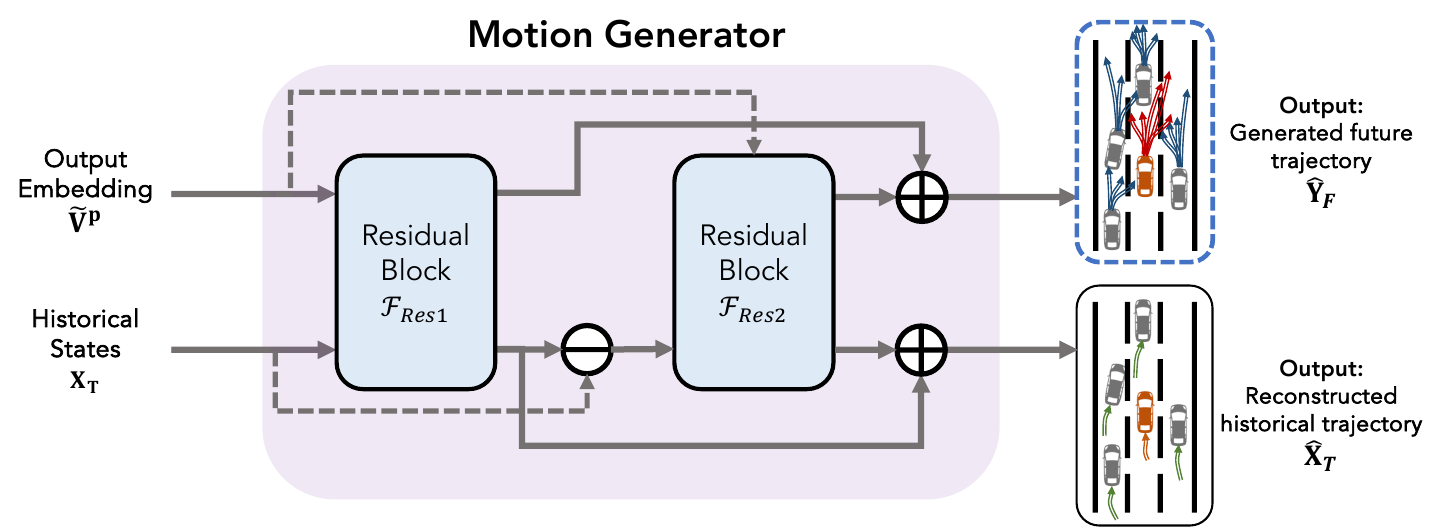}
\caption{Motion Generator.}
\label{fig_motion_generator}
\end{figure}

The Motion Generator fulfills two complementary objectives from the joint embedding \(\widetilde V^p = [\mathbf{Z},\,\widetilde V_T^a]\): (1) reconstructing the past trajectory \(\mathbf{X}_T\) to preserve temporal fidelity and anchor the latent space, and (2) predicting the future trajectory \(\mathbf{Y}_F\) to capture plausible motion under uncertainty.  This dual design prevents mode collapse by ensuring that sampled latent codes \(\mathbf{Z}\) remain grounded in real observations, while the residual structure yields a coarse‐to‐fine refinement of both past and future. Figure \ref{fig_motion_generator} shows the structure of Motion Generator.

The first processing block, $\mathcal{F}_{Res1}$, takes the output embeddings $\mathbf{\tilde V}^{p}$ and the target past trajectory $\mathbf{X}_T$ to generate initial estimates of the future and reconstructed past trajectories $\hat{\mathbf{X}}_{F,1}$ and $\hat{\mathbf{X}}_{T,1}$ respectively:
\begin{equation}
    (\hat{\mathbf{X}}_{T,1},\,\hat{\mathbf{X}}_{F,1})
    = \mathcal{F}_{\mathrm{Res1}}\bigl(\widetilde V^p,\;\mathbf{X}_T\bigr).
\end{equation}
Here, \(\hat{\mathbf{X}}_{T,1}\) reconstructs the historical trajectory, while \(\hat{\mathbf{X}}_{F,1}\) provides an initial future forecast. The second block, \(\mathcal{F}_{\mathrm{Res2}}\), refines both outputs by ingesting the residual error in the past reconstruction,
\(\mathbf{X}_T - \hat{\mathbf{X}}_{T,1}\), along with the same embedding \(\widetilde V^p\):
\begin{equation}
    (\hat{\mathbf{X}}_{T,2},\,\hat{\mathbf{X}}_{F,2})
    = \mathcal{F}_{\mathrm{Res2}}\bigl(\widetilde V^p,\;\mathbf{X}_T - \hat{\mathbf{X}}_{T,1}\bigr).
\end{equation}

Both $\mathcal{F}_{Res1}$ and $\mathcal{F}_{Res2}$ are composed of a GRU encoder for sequence encoding and two MLPs serving as the output header. The final predicted future trajectory $\hat{\mathbf{Y}}_F$ and the reconstructed past trajectory $\hat{\mathbf{X}}_T$ are obtained by summing the respective residuals from both processing blocks:

\begin{equation}
    \hat{\mathbf{Y}}_F = \hat{\mathbf{X}}_{F,1} + \hat{\mathbf{X}}_{F,2}
\end{equation}
\begin{equation}
    \hat{\mathbf{X}}_T = \hat{\mathbf{X}}_{T,1} + \hat{\mathbf{X}}_{T,2}
\end{equation}

Importantly, the Motion Generator is designed to produce not just a single prediction but \(K\) distinct futures for each target vehicle \(i\).  After learning the fused embedding \(\widetilde V^p\), we sample the latent code \(\mathbf{Z}\) from the learned prior distribution \(p(\mathbf{Z}\mid \widetilde V_T^a)\) \(K\) times.  Each sample \(\mathbf{Z}^{(k)}\) is fed through the two‐stage residual decoder to yield a unique future trajectory \(\hat{\mathbf{Y}}_{T+1:T+F}^{(k,i)}\).  Consequently, the Motion Generator outputs 
\begin{equation}
\bigl\{\hat{\mathbf{Y}}_{T+1:T+F}^{(1,i)},\,\dots,\,\hat{\mathbf{Y}}_{T+1:T+F}^{(K,i)}\bigr\},
\end{equation}
providing a diverse set of \(K\) plausible motion hypotheses per target agent. In our experiments, we set \(K=20\) to balance expressivity with computational tractability.

This residual, two‐stage workflow offers several advantages: the first block learns broad motion trends (e.g.\ lane changes, speed profiles), while the second block corrects fine‐grained errors (e.g.\ velocity drift, minor lateral offsets).  Crucially, reconstructing \(\mathbf{X}_T\) anchors the decoder to the true past, preventing the latent sampler from drifting toward unrealistic modes.  The GRU modules enforce temporal smoothness and physically plausible kinematics.  At inference, sampling distinct \(\mathbf{Z}\) from the learned prior yields \(K\) diverse but coherent future trajectories, reflecting both the model’s uncertainty and the fidelity of its learned motion patterns.

\section{Experiments and Results}
\subsection{Data Preparations}
This research leverages two open‐source datasets for the purpose of model training and validation: the Next Generation Simulation (NGSIM) dataset \cite{ngsimdatasetus101,ngsimdatasetus80} and the HighD dataset \cite{highddataset}. The NGSIM dataset provides a comprehensive collection of vehicle trajectory data, capturing activity from the eastbound I‐80 in the San Francisco Bay area and the southbound US 101 in Los Angeles, recorded at 10 Hz. The HighD dataset originates from aerial drone recordings at 25 Hz between 2017 and 2018 near Cologne, Germany, covering 420 m of bidirectional highway and over 110,000 vehicles.
    
\paragraph{\textbf{Data Preprocessing}}
We begin by extracting straight highway segments and retain each dataset’s native sampling (10 Hz for NGSIM; 25 Hz for HighD).  In HighD, isolated missing frames are filled via linear interpolation to preserve the 5 s (50‐frame) prediction horizon; NGSIM requires no resampling.

At each time step \(t\), we form a target vehicle’s interaction group by selecting all vehicles whose centers lie between 100 m behind and 150 m ahead longitudinally, and within one adjacent lane laterally.  For NGSIM, which supplies only the IDs of the single preceding and single following vehicles, we include exactly those two if they satisfy our distance criteria.  For HighD, which provides up to eight surrounding neighbor IDs (preceding, parallel, and following in each of three lanes), we use those IDs and then filter out by distance, yielding at most eight neighbors. In NGSIM we reconstruct analogous contextual neighbors by lane membership and longitudinal ordering: in each adjacent lane we pick the nearest preceding and following vehicles, and the parallel neighbor as the vehicle with the smallest longitudinal gap, followed by the same distance filtering.

Once the group \(\mathcal{V}_i(t)\) is defined, we extract the previous \(T=30\) frames (3 s) of each vehicle’s positions and velocities as the input history \(\mathbf{X}_T\), and the next \(F=50\) frames (5 s) as the ground‐truth future \(\mathbf{Y}_F\).  We split each dataset into 70 \% training and 30 \% testing sets.

\paragraph{\textbf{Maneuver Labeling}}
We restrict behavior enumeration to \emph{lateral} maneuvers rather than full lateral\,$\times$\,longitudinal combinations. To support multi-modal prediction, each vehicle receives a categorical lateral maneuver label at every time step. Following \cite{deo2018multi, chen2022intention}, we consider three lateral classes:
\begin{equation}
\mathcal{M}_{\rm lat}=\{L\ \text{(left lane change)},\;K\ \text{(keep lane)},\;R\ \text{(right lane change)}\}.
\end{equation}
We set the ground-truth label space to $\mathcal{M}\equiv\mathcal{M}_{\rm lat}$ and enforce feasibility with a mask: logits for impossible maneuvers (e.g., $L$ in the leftmost lane) are set to $-\infty$ prior to the softmax, so their probabilities become zero. Importantly, the masked modes in the Agent-Behavior Graph and Hypergraph but are rendered non-participatory by setting all of their incident edges to zero. Thus, the softmax normalizes over feasible options while infeasible modes persist as isolated nodes that do not exchange messages during graph/hypergraph propagation.

\subsection{Training and Evaluation Metrics}

\paragraph{\textbf{Training loss of \texttt{GIRAFFE} }}
To ensure that \texttt{GIRAFFE} learns both accurate trajectories and realistic multi‐agent interactions, we optimize a composite loss:
\begin{equation}
    \mathcal{L}^{\rm PRED}
    = \mathcal{L}_{\rm pred}
    + \mathcal{L}_{\rm int}
    + \mathcal{L}_{\rm fut}\,.
\end{equation}
The term
\begin{equation}
    \mathcal{L}_{\rm pred}
    = \bigl\|\hat{\mathbf{Y}}_F - \mathbf{Y}_F\bigr\|_2^2
\end{equation}
measures the mean‐squared error between the fused multi‐modal forecast \(\hat{\mathbf{Y}}_F\) and the single ground-truth future trajectory \(\mathbf{Y}_F\).  Minimizing \(\mathcal{L}_{\rm pred}\) drives the network to produce trajectories that closely track the actual vehicle motion over the prediction horizon.
To supervise the lateral maneuver classification, we include
\begin{equation}
    \mathcal{L}_{\rm int}
    = -\sum_{i=1}^N\sum_{m\in\{L,K,R\}}
      y_{i,m}\,\log \hat m_{i,m},
\end{equation}
where \(y_{i,m}\) is the one-hot indicator of the true lane‐change mode for agent \(i\), and \(\hat m_{i,m}\) is the softmax probability predicted by the Intention Predictor.  This cross-entropy loss encourages the correct behavior node in \(\mathcal{G}^b\) to be strongly activated, strengthening mode‐specific feature learning.
Finally, we align the learned multi-agent, multi-modal relations with empirical co-occurrence patterns via
\begin{equation}
    \mathcal{L}_{\rm fut}
    = \bigl\|\hat H_F - H_F\bigr\|_F^2,
\end{equation}
where \(H_F\) is the ground-truth adjacency of the Agent–Behavior Graph—extracted from which pairs of agents actually performed simultaneous lane changes—and \(\hat H_F\) is the adjacency inferred from the decoded trajectories and mode probabilities.

\paragraph{\textbf{Training loss of \texttt{RHINO}}} The training loss function for \texttt{RHINO} is also a summation of three components:

\begin{equation} 
    \mathcal{L}^{GEN} =  \mathcal{L}_{elbo} + \mathcal{L}_{recon} + \mathcal{L}_{var} 
\end{equation}

The first term \(\mathcal{L}_{\mathrm{elbo}}\) is the Evidence Lower Bound \cite{pagnoni2018conditional} on the conditional log‐likelihood of the true future \(\mathbf{Y}_F\), and explicitly trades off reconstruction fidelity against uncertainty regularization:
\begin{equation} 
    \mathcal{L}_{\mathrm{elbo}} 
    = -\,\mathbb{E}_{q_\phi(\mathbf{Z}\mid \widetilde V_T,\widetilde V_F)}\bigl[\log p_\theta(\mathbf{Y}_F\mid \mathbf{Z},\widetilde V_T)\bigr] \\
     + \mathrm{KL}\!\Bigl[q_\phi(\mathbf{Z}\mid \widetilde V_T,\widetilde V_F)
      \,\Big\|\,p_\psi(\mathbf{Z}\mid \widetilde V_T)\Bigr].
\end{equation}
The EBLO loss measures how well sampled latent codes \(\mathbf{Z}\) rebuild the true future—driving the network to capture the dominant motion patterns—while the KL‐divergence term forces the learned posterior distribution to remain close to the prior \(p_\psi(\mathbf{Z}\mid \widetilde V_T)\).  This regularization prevents the model from becoming over‐confident in a narrow set of trajectories and encourages a spread of plausible futures, thereby quantifying predictive uncertainty.  
In practice we implement this as
\begin{equation} 
    \mathcal{L}_{\mathrm{elbo}}
    = \alpha\,\bigl\|\hat{\mathbf{Y}}_F - \mathbf{Y}_F\bigr\|_2^2
    + \beta\,\mathrm{KL}\!\Bigl[\mathcal{N}(\mu_q,\mathrm{Diag}(\sigma_q^2))
      \;\Big\|\;\mathcal{N}(0,\lambda I)\Bigr].
\end{equation}

The second component, $\mathcal{L}_{recon}$, represents the Historical Trajectory Reconstruction loss, which measures how accurately the reconstructed historical trajectories match the true historical data:

\begin{equation}
    \mathcal{L}_{recon} = \lambda \| \hat{\mathbf{X}}_T - \mathbf{X}_T \|_2^2
\end{equation}

The final component, $\mathcal{L}_{var}$, is the Variety loss, inspired by Social-GAN \cite{gupta2018social}. This loss encourages diversity in the predicted future trajectories by minimizing the error across multiple sampled future trajectories:

\begin{equation} 
    \mathcal{L}_{var} = \min_k \| \hat{\mathbf{Y}}_F^{(k)} - \mathbf{Y}_F \|_2^2 
\end{equation}

Table \ref{tab:parameter} presents the hyperparameter configurations used for the network architecture and the training process in the proposed framework.

\begin{table}[!ht]
\footnotesize
\caption{Hyperparameter Settings}\label{tab:parameter}
\centering
\begin{tabular}{>{\centering\arraybackslash}m{3cm} >{\centering\arraybackslash}m{1.2cm} |>{\centering\arraybackslash}m{3cm} >{\centering\arraybackslash}m{1.2cm}}
\toprule
Parameter & Value & Parameter & Value\\
\midrule
$T$ & 30  & learning rate & 0.001\\
$F$ & 50 & decaying factor & 0.6\\
neuron \# of MLPs & 128 & $\alpha$ & 1\\
$s$ & 3 & $\beta$ & 0.8\\         
$\{J^{(s)}\}$ & $\{2,3,5\}$ & $\lambda$ & 0.5\\
\bottomrule
\end{tabular}
\end{table}

\paragraph{\textbf{Evaluation metrics}} To ascertain the predictive accuracy of the model, we employ the Root Mean Square Error (RMSE) as the evaluative criterion. This metric quantitatively measures the deviation between the predicted position, expressed as $(\hat y_{f,lat}^l, \hat y_{f,lon}^l)$, and the ground truth position, indicated by $(y_{f, lat}^l,y_{f, lon}^l)$ for all time step $f$ within the predictive horizon prediction horizon $[T+1, T+F]$.
\begin{equation}
    RMSE= \sqrt{\frac{1}{LF}\sum_{l=1}^L \sum_{f=T+1}^{T+F} \left((\hat y_{f,lat}^l - y_{f,lat}^l)^2 + (\hat y_{f,lon}^l - y_{lon}^l)^2 \right)}
\end{equation}
where the superscript $l$ denotes the $l$-th test sample from the aggregate test sample set with length $L$.

\subsection{Results of Trajectory Generation}
\begin{figure*}[!ht]
\centering
\includegraphics[width=1\textwidth]{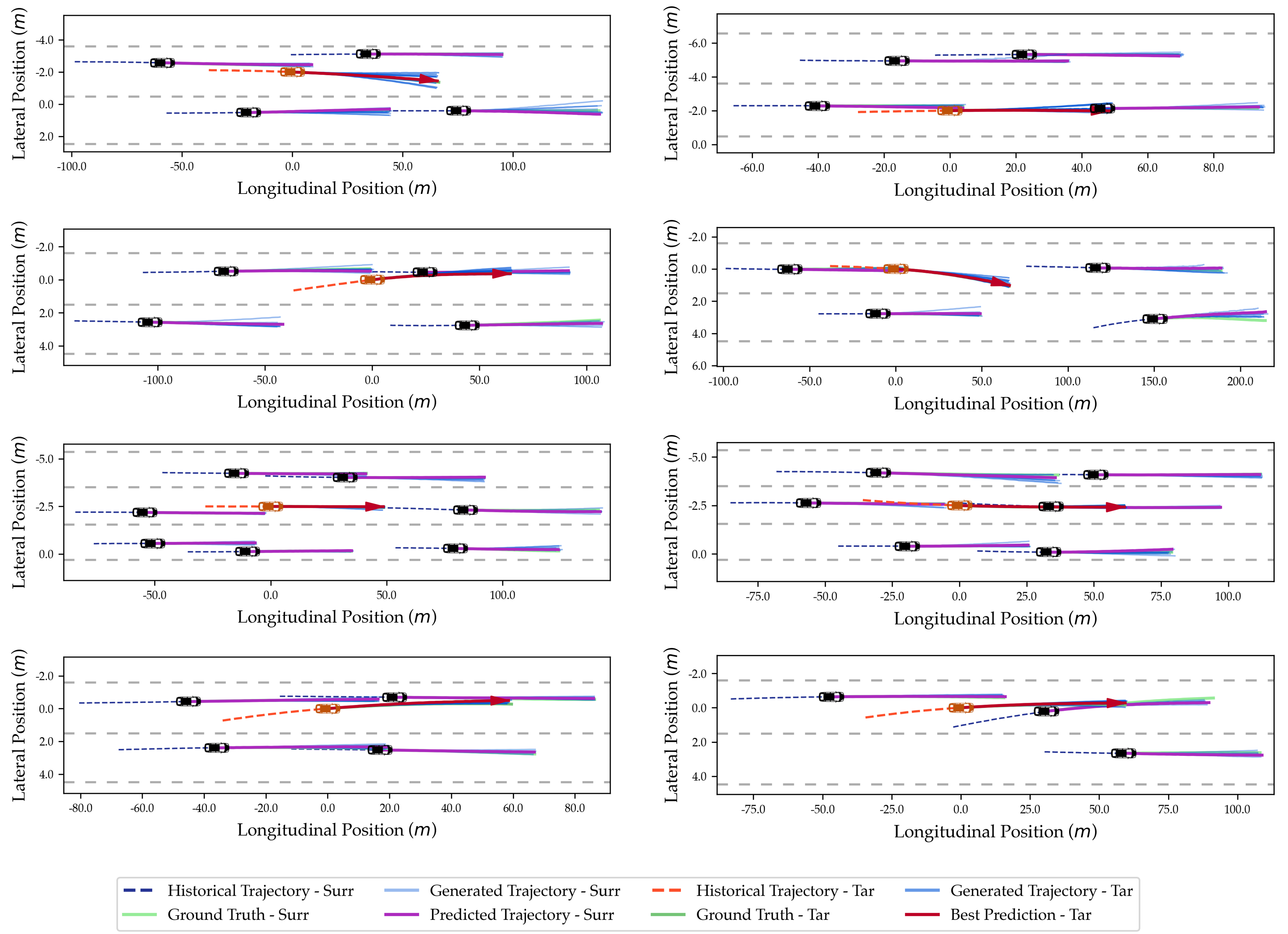}
\caption{Trajectory generation results in highway scenarios.}
\label{fig_plan_gen}
\end{figure*}

\begin{figure*}[!ht]
\centering
\includegraphics[width=0.7\textwidth]{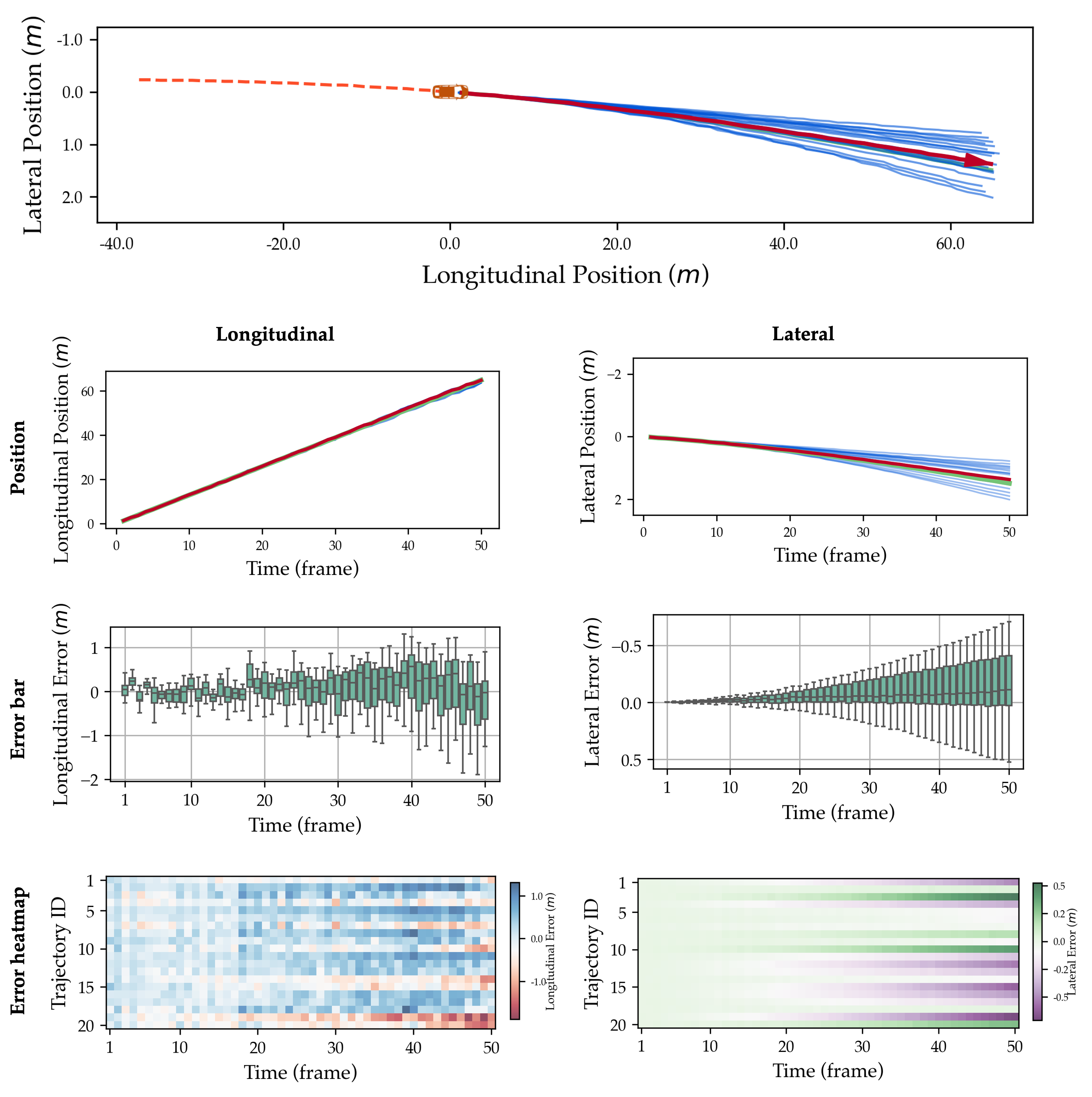}
\caption{Longitudinal and lateral trajectory generation error analysis.}
\label{fig_gen_err1}
\end{figure*}

The experimental results for trajectory generation of the $K$ trajectories using the HighD dataset are presented in Figure \ref{fig_plan_gen}. As can be found that, \texttt{RHINO} demonstrates strong generative capabilities, effectively producing plausible motion in a dynamic interactive traffic environment. To provide a more quantitative analysis, trajectory generation inaccuracies are illustrated in Figure \ref{fig_gen_err1}. The generated longitudinal and lateral trajectories, along with the error box plots and heatmaps, are displayed. The box plot reveals that errors in both axes increase with the prediction time step by the natural of error propagation. However, the errors remain within an acceptable range, indicating decent model performance, which demonstrates high precision in trajectory generation. Notably, the model maintains a lower error margin for shorter prediction horizons, which is critical for short-term planning and reactive maneuvers in dynamic traffic environment.

\begin{figure*}[!ht]
\centering
\includegraphics[width=0.9\textwidth]{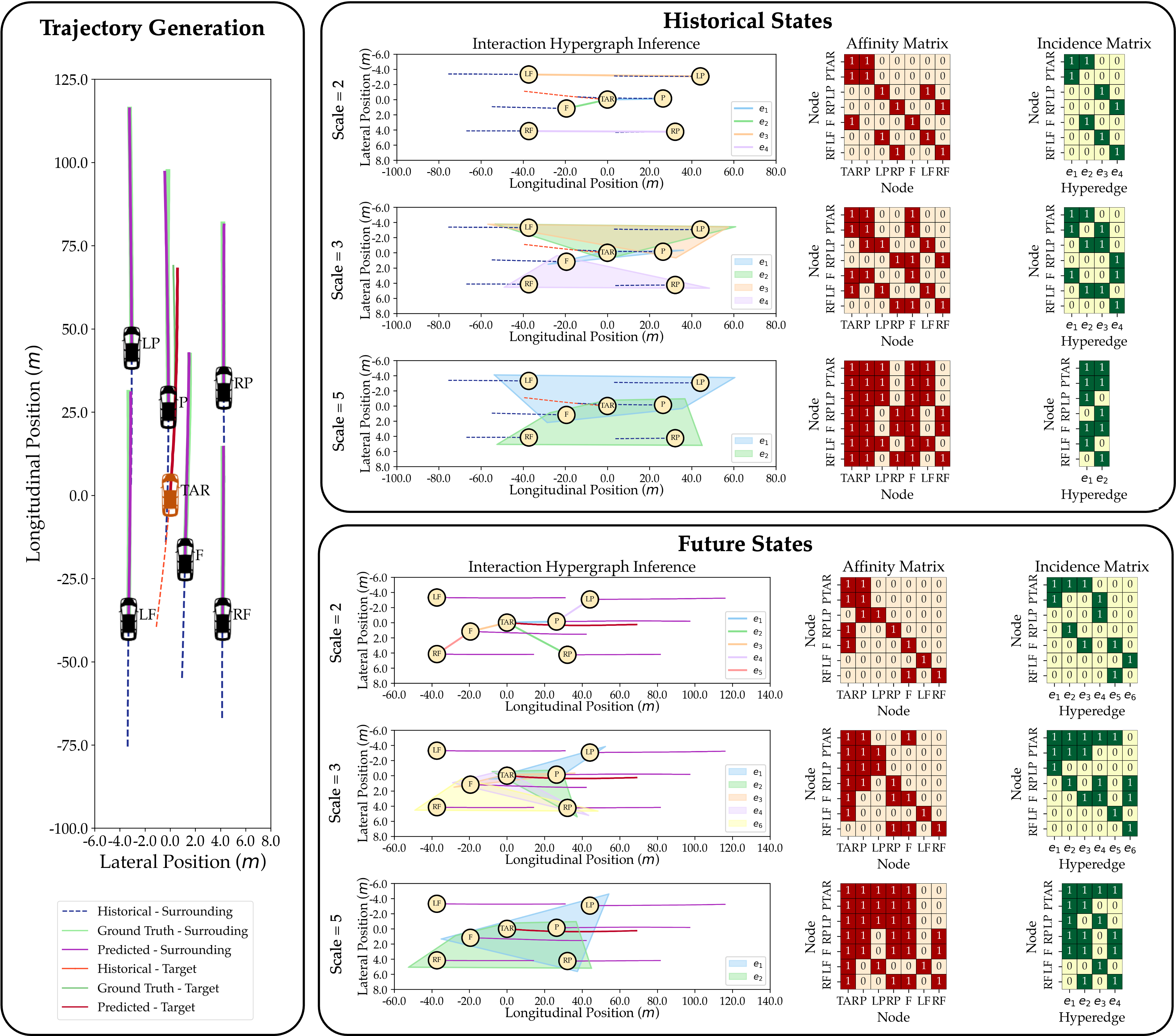}
\caption{Trajectory generation with hypergraph inference.}
\label{fig_hypergraph_inference1}
\end{figure*}

The experiment focused on predicting vehicle trajectories within mixed traffic environments on highways by employing hypergraph inference to model group-based interactions. Through the application of hypergraph models at varying scales \( s = 2, 3, 5 \), the experiment captured the evolution of multi-vehicle interactions across both historical and future horizons. The figures depict these dynamics through three distinct columns: the first column presents vehicle trajectories and the corresponding hyperedges, visualized as polygons that encapsulate groups of interacting vehicles. The second column illustrates the affinity matrix, where both rows and columns represent vehicles, and the strength of their relationships is indicated by the matrix values. The third column shows the incidence matrix, detailing the relationship between nodes and hyperedges, with each column representing a hyperedge and each vehicle's involvement in that hyperedge marked by a 1 in the corresponding row.

The hypergraph-based approach is particularly effective in modeling complex, higher-order interactions that are beyond the scope of traditional pairwise models. By forming hyperedges that encompass multiple vehicles, the model captures the collective influence that a group’s behavior exerts on an individual vehicle. For instance, in the scenario shown in Figure \ref{fig_hypergraph_inference1}, at scale \( s = 5 \), when the target vehicle TAR initiates a lane change, the hypergraph reflects the interaction not only with a single neighboring vehicle but also with multiple surrounding vehicles, such as following vehicle F,  preceding vehicle P, and preceding vehicle RP in the right lane. More examples are illustrated in \ref{app_exp}. This capability to model group-wise interactions across different scales is essential for accurately predicting vehicle trajectories in congested highway environments, where the actions of one vehicle can trigger ripple effects that influence an entire group. The hypergraph’s dynamic formation of hyperedges ensures that predicted trajectories remain adaptable and responsive to broader traffic conditions.

\subsection{Comparisons and Ablation Study}

To evaluate the privileges of our proposed method, the state of art methods (i.e., Social-LSTM (S-LSTM) \cite{alahi2016social}, Convolutional Social-LSTM (CS-LSTM) \cite{deo2018convolutional}, Planning-informed prediction (PiP) \cite{song2020pip}, Graph-based Interaction-aware Trajectory Prediction (GRIP) \cite{li2019grip}, Spatial-temporal dynamic attention network (STDAN) \cite{chen2022intention}, WSiP \cite{wang2023wsip}, and GaVa \cite{liao2024human} are compared.

\begin{figure}[!ht]
\centering
\includegraphics[width=1.0\textwidth]{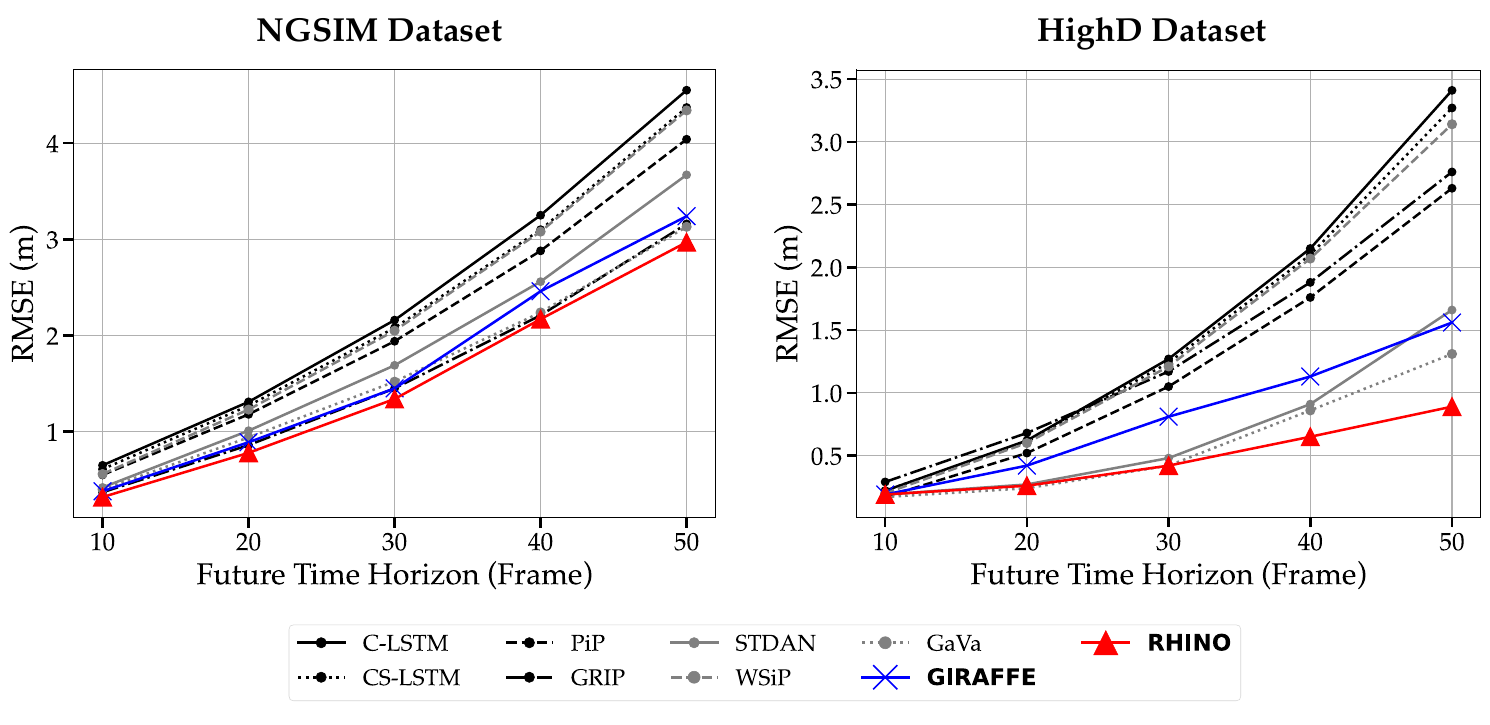}
\caption{Prediction error obtained by different models in RMSE on NGSIM dataset (left) and HighD dataset (right).}
\label{fig_result_comp}
\end{figure}

The compared results presented in Table \ref{tab:comparison} and Figure \ref{fig_result_comp}. As can be found that, the proposed framework demonstrates good performance with respect to the RMSE across a prediction horizon of 50 frames when compared with existing baseline models. It exhibits a reduced loss in comparison to C-LSTM, CS-LSTM, PiP, GRIP, and WSiP. These outcomes suggest that the proposed model effectively captures salient features pertinent to long-term predictions. In summary, the proposed framework outperforms baseline models on the HighD dataset and delivers commendable performance on the NGSIM dataset.

Since the \texttt{RHINO} adopts the \texttt{GIRAFFE}, we further compare the trajectory generation capability of \texttt{RHINO} with our previous work \cite{wu2023graph} and its enhanced version \texttt{GIRAFFE}. Both \texttt{RHINO} model and the enhanced \texttt{GIRAFFE} model consistently outperform the baseline models, demonstrating superior performance in various metrics. This suggests that our proposed approaches effectively address the limitations present in prevailing models by robustly capturing complex interactions.


\begin{table}[!ht]
    \footnotesize
    \caption{Prediction Error Obtained by Different Models in RMSE (m)}\label{tab:comparison}
    \centering
    \begin{tabular}{>{\centering\arraybackslash}m{1.2cm} >{\centering\arraybackslash}m{1.0cm} >{\centering\arraybackslash}m{0.8cm} >{\centering\arraybackslash}m{0.8cm} >{\centering\arraybackslash}m{0.8cm} >
    {\centering\arraybackslash}m{0.8cm} >{\centering\arraybackslash}m{1.0cm} > 
    {\centering\arraybackslash}m{1.0cm} > {\centering\arraybackslash}m{1.2cm} 
    |>{\centering\arraybackslash}m{1.2cm} >{\centering\arraybackslash}m{1.2cm}}
    \toprule
    Dataset & Horizon (Frame) & S-LSTM & CS-LSTM & PiP & GRIP & STDAN & WSiP & GaVa & \texttt{GIRAFFE} & \texttt{RHINO}\\
    \midrule
    \multirow{5}{*}{NGSIM} & 10 & 0.65 &0.61 & 0.55 & \underline{0.37} & 0.42 & 0.56 & 0.40 & 0.38 & \textbf{0.32}\\
     & 20 & 1.31 & 1.27 & 1.18 & \underline{0.86} & 1.01 & 1.23 & 0.94 & 0.89 & \textbf{0.78}\\
     & 30 & 2.16 & 2.08 & 1.94 & \underline{1.45} & 1.69 & 2.05 & 1.52 & \underline{1.45} & \textbf{1.34}\\
     & 40 & 3.25 & 3.10 & 2.88 & \underline{2.21} & 2.56 & 3.08 & 2.24 & 2.46 & \textbf{2.17}\\
     & 50 & 4.55 & 4.37 & 4.04 & 3.16 & 3.67 & 4.34 & \underline{3.13} & 3.24 & \textbf{2.97}\\
    \midrule
    \multirow{5}{*}{HighD} & 10 & 0.22 & 0.22 & \textbf{0.17} & 0.29 & \underline{0.19} & 0.20 & \textbf{0.17} & \underline{0.19} & \underline{0.19}\\
     & 20 & 0.62 & 0.61 & 0.52 & 0.68 & 0.27 & 0.60 & \textbf{0.24} & 0.42 & \underline{0.26} \\
     & 30 & 1.27 & 1.24 & 1.05 & 1.17 & \underline{0.48} & 1.21  & \textbf{0.42} & 0.81 & \textbf{0.42}\\
     & 40 & 2.15 & 2.10 & 1.76 & 1.88 & 0.91 & 2.07 & \underline{0.86} & 1.13 & \textbf{0.65}\\
     & 50 & 3.41 & 3.27 & 2.63 & 2.76 & 1.66 & 3.14 & \underline{1.31} & 1.56 & \textbf{0.89}\\
    \bottomrule
    \end{tabular}
\end{table}

Ablation study is conducted to provide more insights into the performance of our \texttt{RHINO} model, especially the impact of different components on the prediction performance by disabling the corresponding component from the entire \texttt{RHINO}. In particular, we consider the following four variants:
\begin{itemize}
    \item \textbf{\texttt{RHINO w/o HG}} (hypergraph) variant does not use the multi-scale hypergraphs representation but only adopts the pair-wise connected graph representations in the Hypergraph Relational Encoder. 
    \item \textbf{\texttt{RHINO w/o MM}} (multi-modal) variant does not adopt the multi-agent multi-modal trajectory prediction results and only use the single predicted future states for each agent as the input of the \texttt{RHINO}. 
    \item \textbf{\texttt{RHINO w/o PDL}} (posterior distribution learner) variant skips the Posterior Distribution Learner and directly input the graph embedding into the Motion Generator.
\end{itemize}

\begin{table}[!t]
\caption{Ablation Test Results of \texttt{RHINO} in RMSE  (m)}\label{table_abalation_test}
\centering
\small
\begin{tabular}{>{\centering\arraybackslash}m{1.5cm} >
{\centering\arraybackslash}m{1.5cm} >{\centering\arraybackslash}m{1.5cm} >{\centering\arraybackslash}m{1.5cm} >{\centering\arraybackslash}m{1.5cm}}
\hline\hline
Horizon (Frame) &  RHINO w/o HG & RHINO w/o MM & RHINO w/o PDL & RHINO \\
\hline
10 & 0.21 & 0.22 & 0.24 & \textbf{0.19} \\
20 & 0.31 & 0.37 & 0.42 & \textbf{0.26} \\
30 & 0.68 & 0.73 & 0.80 & \textbf{0.42} \\
40 & 0.97 & 1.06 & 1.18 & \textbf{0.65} \\
50 & 1.25 & 1.34 & 1.57 & \textbf{0.89} \\
\hline\hline
\end{tabular}
\end{table}

\begin{figure*}[!t]
\centering
\includegraphics[width=0.6\textwidth]{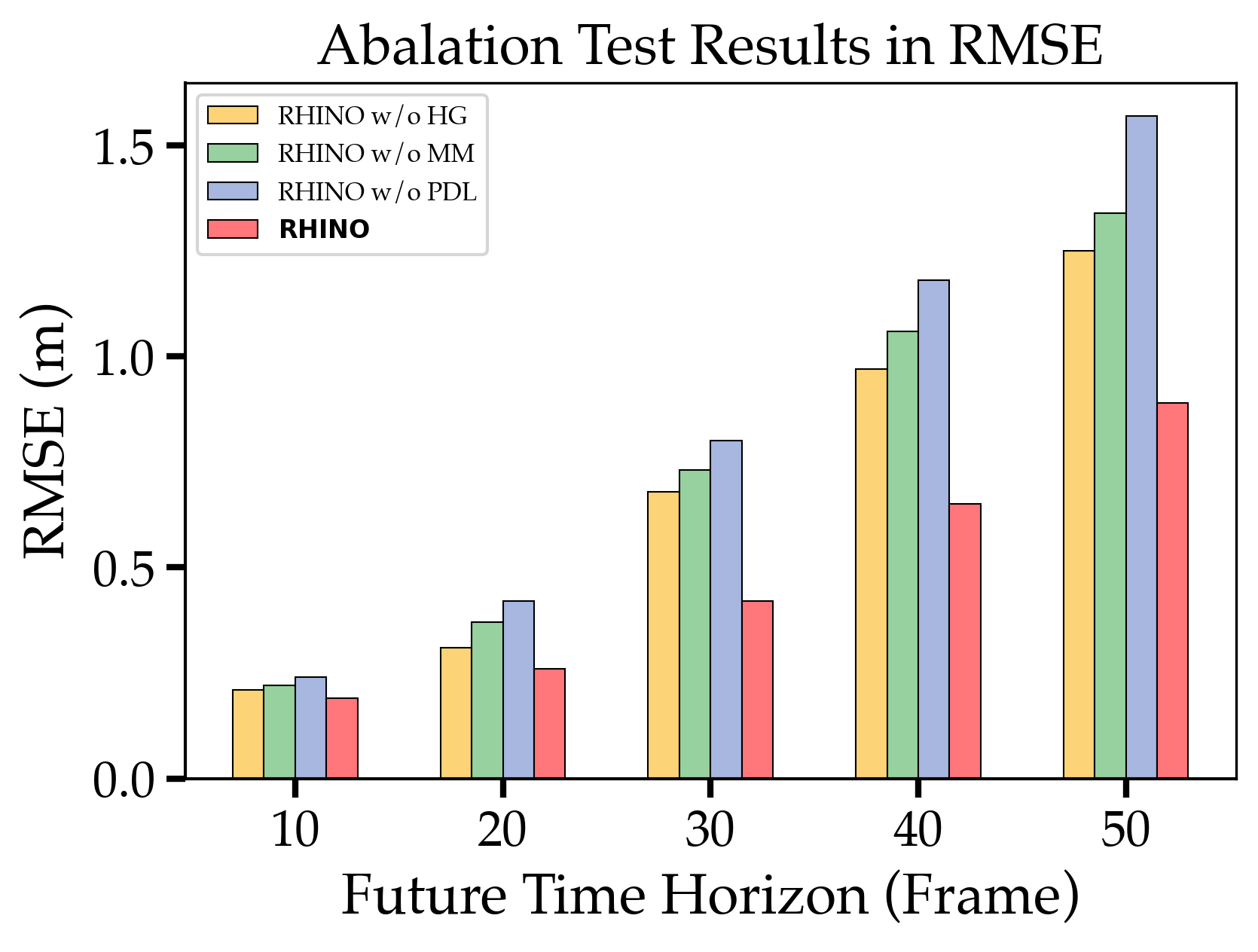}
\caption{Ablation Study of \texttt{RHINO}.}
\label{fig_abalation_test}
\end{figure*}

An investigation into the effects of model design variations, as presented in Table \ref{table_abalation_test} and Figure \ref{fig_abalation_test}. The removal of various components from \texttt{RHINO} invariably leads to performance degradation to varying degrees. Compared to the full \texttt{RHINO}, omitting the multi-scale hypergraphs results in an evident increase in prediction error across the prediction horizon, indicating that modeling and reasoning group-wise interactions using hypergraphs, rather than solely pair-wise interactions, enhances prediction accuracy which underscores the necessity of hypergraphs. Further, excluding the multi-agent multi-modal trajectory prediction input leads to a more substantial degradation in performance, highlighting the importance of incorporating multi-modal motion states and discussing the corresponding group-wise interactions among multiple driving behaviors of multiple agents. Lastly, the absence of the Posterior Distribution Learner module emphasizes its critical role in handling the stochasticity of each agent’s behavior. All these experiments justify the  effectiveness of the full model.

\section{Conclusions}
In this study, we proposed a hypergraph enabled multi-modal probabilistic motion prediction framework with reasonings. This framework consists of two main components: \texttt{GIRAFFE} and \texttt{RHINO}. \texttt{GIRAFFE} focuses on predicting the interactive vehicular trajectories considering modalities. Based on that, \texttt{RHINO}, leveraging the flexibility and strengths on modeling the group-wise interactions, facilitate relational reasoning among vehicles and multi-modalities to render plausible vehicles trajectories. The framework extends traditional interaction models by introducing an agent-behavior hypergraph. This approach better aligns with traffic physics while being grounded in the mathematical rigor of hypergraph theory. Further, the approach employs representation learning to enable explicit interaction relational reasoning. This involves considering future relations and interactions and learning the posterior distribution to handle the stochasticity of  behavior for each vehicle.  As a result, the framework excels in capturing high-dimensional, group-wise interactions across various behavioral modalities. The framework is tested using the NGSIM and HighD datasets. The results show that the proposed framework effectively models the interactions among groups of vehicles and their corresponding multi-modal behaviors. Comparative studies demonstrate that the framework outperforms prevailing algorithms in prediction accuracy. To further validate the effectiveness of each component, ablation studies were conducted, revealing that the full model performs best.

While the hypergraph design increases expressivity, it also raises computational demand: topology must be inferred at multiple granularities, message passing operates over many-to-many incidences, and the probabilistic decoder produces multiple trajectory samples. In practice, we keep the hypergraph compact (few scales with moderate group sizes), sparsify aggressively (spatial gating and pruning low-affinity hyperedges), and reuse decoder computation across samples to control runtime. A related challenge is hyperparameter tuning: performance can be sensitive to the number of scales, per-scale group size, embedding width, message-passing depth, and the number of sampled futures. We provide a practical operating region and sensitivity evidence in the appendix; denser or more heterogeneous scenes may warrant light retuning.

Several promising extensions remain. We will broaden evaluation beyond freeways to highly interactive urban regimes—arterials, signaled intersections, and roundabouts—and to widely used public benchmarks (e.g., Waymo Open Motion, nuScenes) \cite{sun2020scalability, caesar2020nuscenes}. To reduce domain gaps, we will standardize map projections and derive lane polylines from lane markings when HD maps are incomplete, enabling consistent inputs across datasets. The state representation will be enriched with HD lane geometry/topology, vehicle attributes, signal phase and timing (SPaT), and V2V/V2I messages \cite{wu2025digital}; we will also encode environmental context (e.g., precipitation proxies, friction surrogates) to modulate motion priors under adverse conditions~\cite{zhang2025virtual}. To keep inference tractable at scale, we plan to employ compact, aggressively sparse hypergraph constructions alongside computation sharing in the decoder. Finally, we aim to improve interpretability via hyperedge-level saliency maps, discovery of prototypical interaction motifs, and language-grounded (LLM) rationales \cite{wu2025v2x}, together with per-scenario breakdowns and calibrated uncertainty reporting, yielding practitioner-oriented explanations of group-wise behavior and model confidence.

\section*{\textbf{Acknowledgement:}}
This research is funded by Federal Highway Administration (FHWA) Exploratory Advanced Research 693JJ323C000010. The results do not reflect FHWA's opinions.



 \bibliographystyle{elsarticle-num} 
 \bibliography{ref}



\newpage
\appendix
\input{appendix}

\end{document}

%% file: appendix.tex
\section{Experiment results of trajectory generation with hypergraph inference} \label{app_exp}

\begin{figure*}[!ht]
\centering
\includegraphics[width=0.9\textwidth]{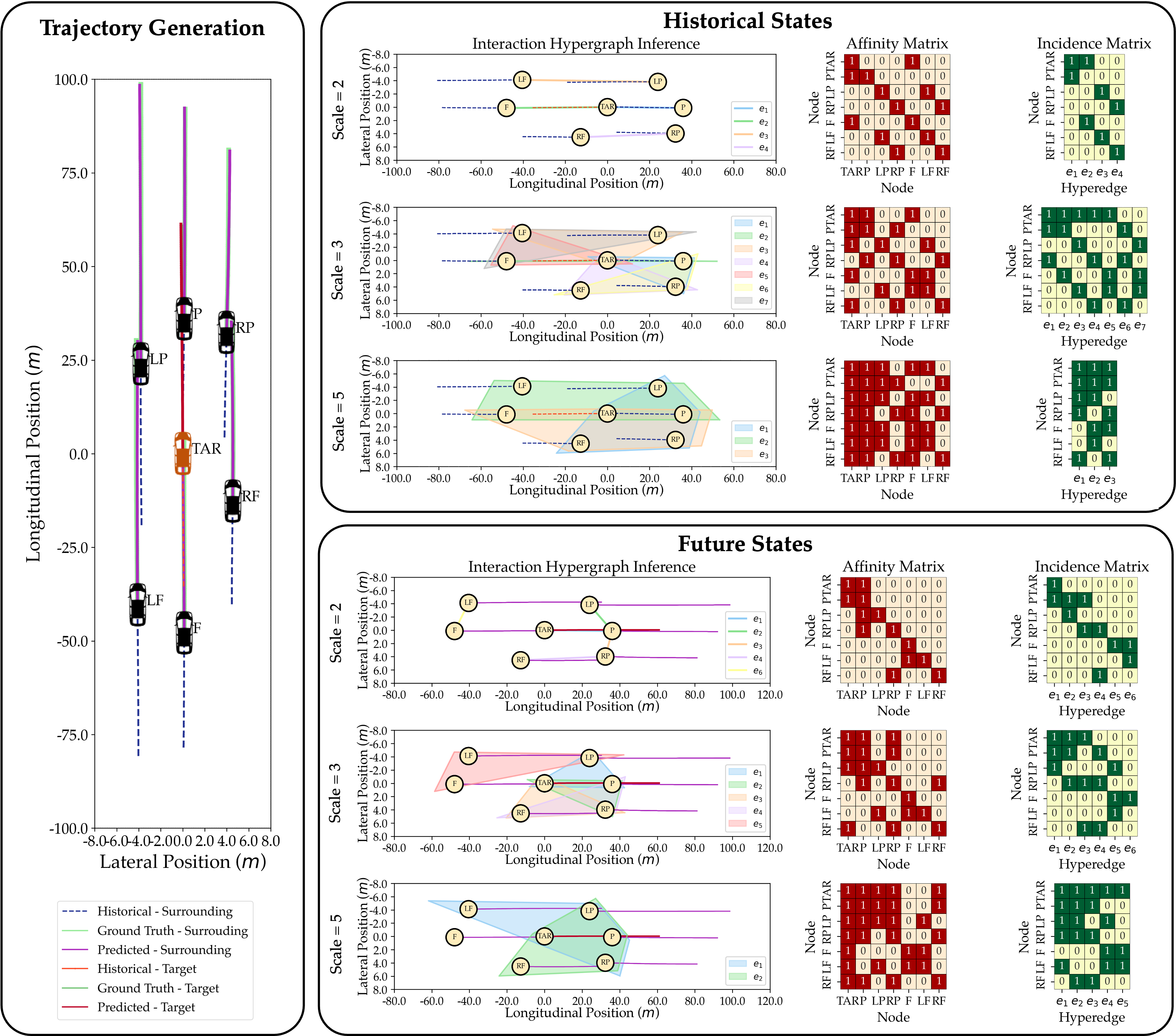}
\caption{Scenario 2 of Trajectory generation with hypergraph inference.}
\label{fig_hypergraph_inference2}
\end{figure*}

In Scenario 2, at scale $s = 5$ (Figure \ref{fig_hypergraph_inference2}), the target vehicle TAR forms a strong historical interaction with multiple vehicles, including the preceding vehicle P, preceding vehicle LP in the left lane, and following vehicles LF and F in the left lane. However, in the future states, vehicle LF exits the interaction group, while the preceding vehicle RP in the left lane joins the group as the target vehicle accelerates toward RF.

\begin{figure*}[!ht]
\centering
\includegraphics[width=0.9\textwidth]{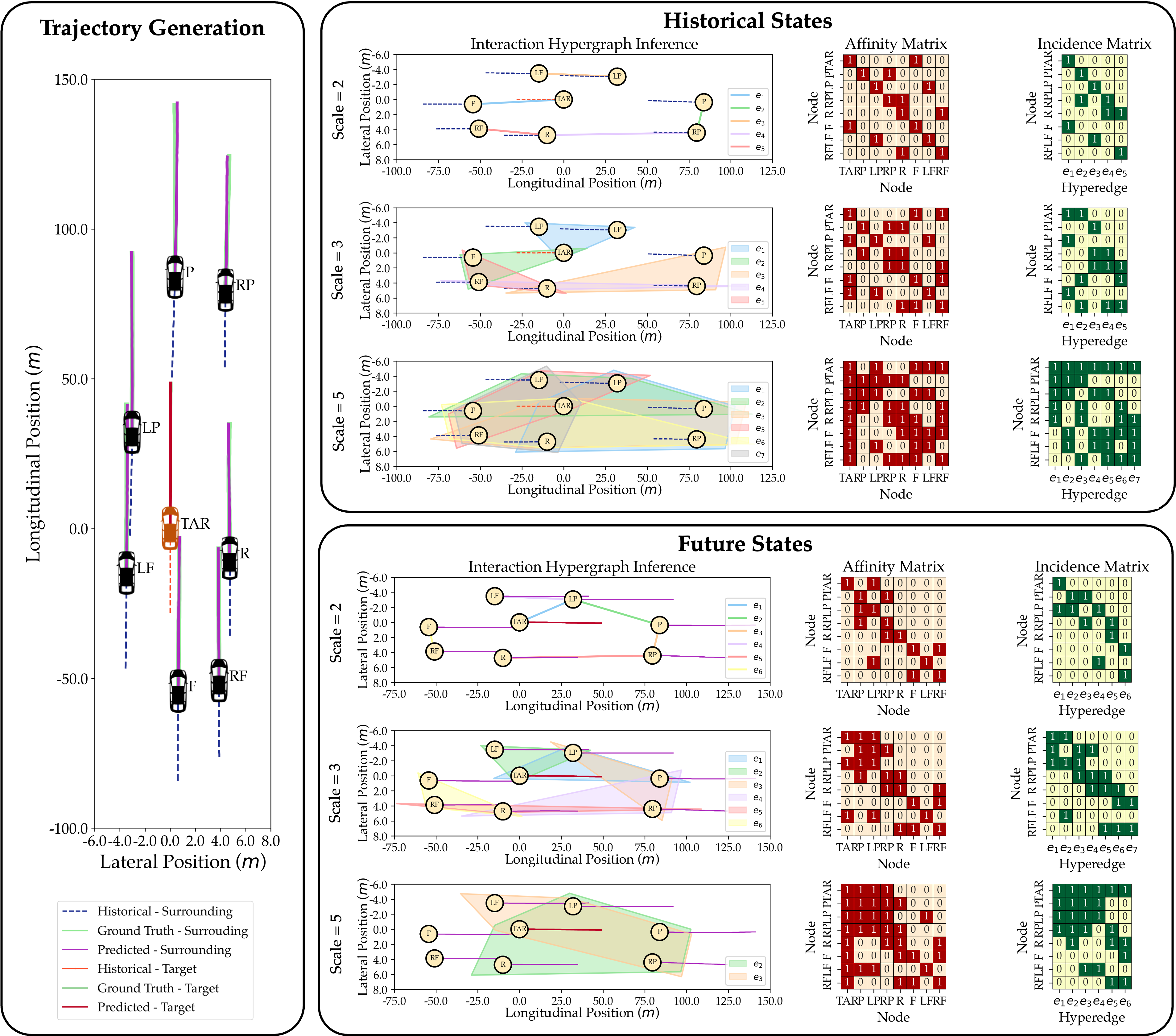}
\caption{Scenario 3 of Trajectory generation with hypergraph inference.}
\label{fig_hypergraph_inference3}
\end{figure*}

In Scenario 3, at scale $s = 3$ (Figure \ref{fig_hypergraph_inference3}), the target vehicle TAR forms a weak historical interaction with the preceding vehicles P and LP in the left lane. As traffic conditions change over time, these interactions strengthen in the future, highlighting the need to account for dynamic shifts when predicting future states, as historical data alone may not capture the evolving complexity of vehicle interactions. These findings emphasize the importance of adaptive models in predicting multi-agent traffic dynamics.

\section{Hyperparameter Sensitivity Analysis}

In this appendix, we present a comprehensive evaluation of RHINO’s robustness to five key hyperparameters. Each study was conducted on the HighD dataset by varying one parameter at a time, while holding all others at their default settings:
\[
  S=3,\quad
  \{J^{(0)},J^{(1)},J^{(2)}\}=\{2,3,5\},\quad
  \text{Latent MLP layers}=1,\quad
  K=20,\quad
  d=128.
\]
We report lateral RMSE (in meters) at prediction horizons of 10, 20, 30, 40, and 50 frames. Across all experiments, RHINO’s performance remains accurate and stable, demonstrating its insensitivity to modest hyperparameter perturbations.

\begin{figure}[!ht]
      \centering
      \includegraphics[width=\textwidth]{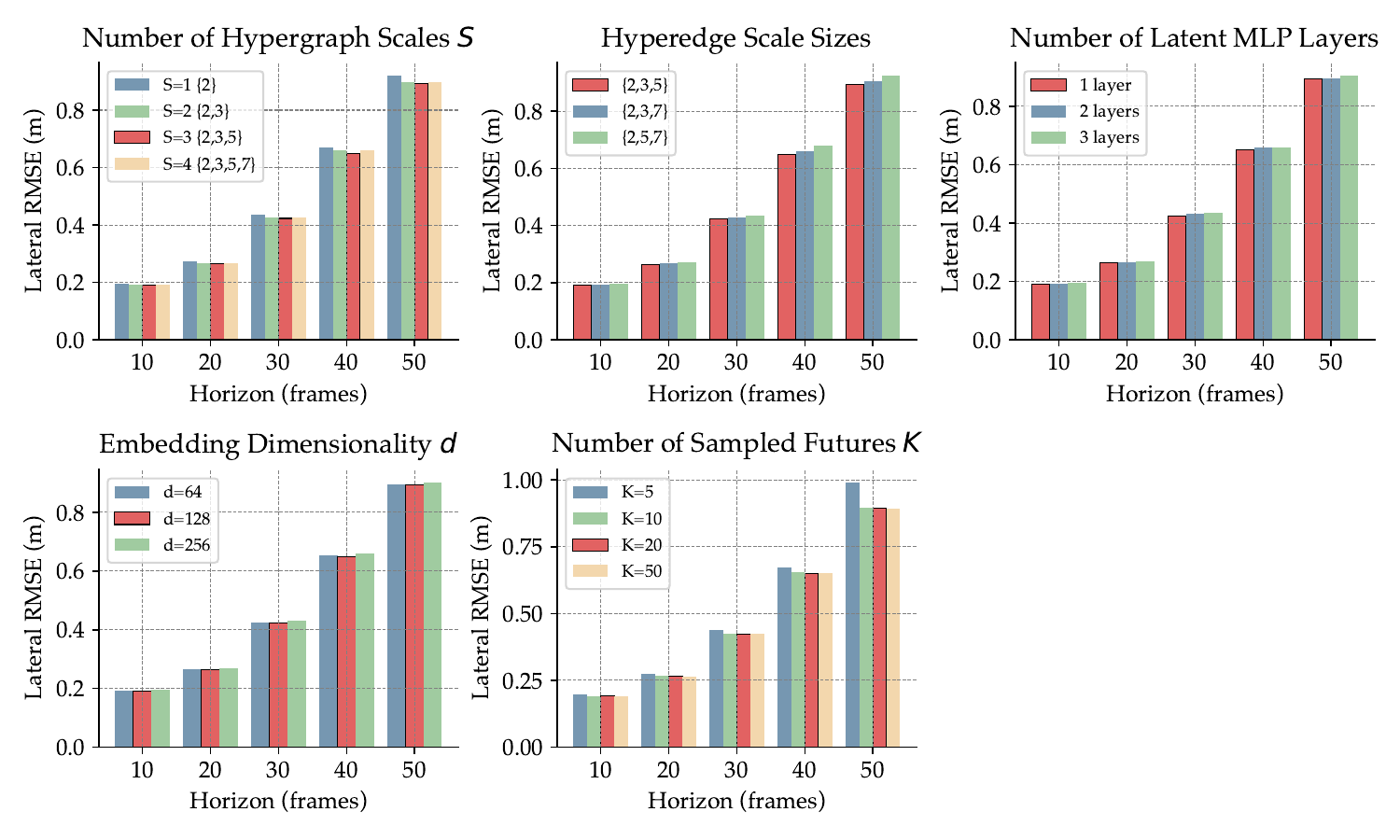}
      \caption{Sensitivity analysis results.}
      \label{fig:sensitivity}
    \end{figure}

Figure~\ref{fig:sensitivity} summarizes RHINO’s hyperparameter sensitivity: across all sweeps, our chosen defaults (highlighted in red bars) achieve near-optimal lateral RMSE and demonstrates the robustness of the model settings.

\subsection{Number of Hypergraph Scales \(S\)}

Traffic behaviors emerge from both pairwise interactions and larger vehicle clusters. We compare:
\[
  S=1:\{2\},\quad
  S=2:\{2,3\},\quad
  S=3:\{2,3,5\},\quad
  S=4:\{2,3,5,7\}.
\]
Table~\ref{tab:sensitivity_scales} shows that increasing \(S\) reduces RMSE through \(S=3\), with only marginal gains at \(S=4\). This indicates that hyperedges up to size five capture the bulk of group-wise effects; larger groups are relatively rare and contribute little extra predictive power. The improvement from \(S=2\) to \(S=3\) grows with horizon, highlighting the importance of medium-sized clusters for longer-term forecasting.

\begin{table}[h!]
  \footnotesize
  \centering
  \caption{Sensitivity to Number of Hypergraph Scales \(S\).}
  \label{tab:sensitivity_scales}
  \begin{tabular}{lcccc}
  \toprule
  Horizon & \(S=1\,\{2\}\) & \(S=2\,\{2,3\}\) & \(\mathbf{S=3\,\{2,3,5\}}\) & \(S=4\,\{2,3,5,7\}\) \\
  \midrule
  10 & 0.197 & 0.191 & \textbf{0.191} & 0.192 \\
  20 & 0.275 & 0.266 & \textbf{0.264} & 0.266 \\
  30 & 0.435 & 0.427 & \textbf{0.423} & 0.426 \\
  40 & 0.670 & 0.659 & \textbf{0.650} & 0.659 \\
  50 & 0.920 & 0.896 & \textbf{0.893} & 0.897 \\
  \bottomrule
  \end{tabular}
\end{table}

\subsection{Hyperedge Scale Sizes \(\{J^{(0)},J^{(1)},J^{(2)}\}\)}

Holding \(S=3\), we vary the maximum group sizes per scale. Table~\ref{tab:sensitivity_sizes} shows that our default \(\{2,3,5\}\) achieves the lowest RMSE, confirming that moderate platoon sizes best reflect highway dynamics. Larger hyperedges add complexity with diminishing returns.

\begin{table}[h!]
  \footnotesize
  \centering
  \caption{Sensitivity to Hyperedge Scale Sizes \(\{J^{(0)},J^{(1)},J^{(2)}\}\).}
  \label{tab:sensitivity_sizes}
  \begin{tabular}{lccc}
  \toprule
  Horizon & \(\mathbf{\{2,3,5\}}\) & \(\{2,3,7\}\) & \(\{2,5,7\}\) \\
  \midrule
  10 & \textbf{0.191} & 0.193 & 0.195 \\
  20 & \textbf{0.264} & 0.268 & 0.272 \\
  30 & \textbf{0.423} & 0.428 & 0.435 \\
  40 & \textbf{0.650} & 0.660 & 0.679 \\
  50 & \textbf{0.893} & 0.905 & 0.924 \\
  \bottomrule
  \end{tabular}
\end{table}

\subsection{Number of Latent MLP Layers}

We evaluate 1 (default), 2, and 3 hidden layers in the Posterior Distribution Learner. As Table~\ref{tab:sensitivity_layers} shows, a single layer suffices to capture trajectory uncertainty; deeper encoders slightly degrade performance due to over-parameterization.

\begin{table}[h!]
  \footnotesize
  \centering
  \caption{Sensitivity to Number of Latent MLP Layers.}
  \label{tab:sensitivity_layers}
  \begin{tabular}{lccc}
  \toprule
  Horizon & \(\mathbf{1\text{ layer}}\) & \(2\) layers & \(3\) layers \\
  \midrule
  10 & \textbf{0.191} & 0.193 & 0.195 \\
  20 & \textbf{0.264} & 0.265 & 0.270 \\
  30 & \textbf{0.423} & 0.432 & 0.435 \\
  40 & \textbf{0.650} & 0.658 & 0.660 \\
  50 & \textbf{0.893} & 0.896 & 0.905 \\
  \bottomrule
  \end{tabular}
\end{table}

\subsection{Embedding Dimensionality \(d\)}

We test \(d\in\{64,128,256\}\). Table~\ref{tab:sensitivity_dim} confirms that \(d=128\) strikes the best balance: lower dimensions lack capacity, while larger ones introduce redundant parameters without accuracy gains.

\begin{table}[h!]
  \footnotesize
  \centering
  \caption{Sensitivity to Embedding Dimensionality \(d\).}
  \label{tab:sensitivity_dim}
  \begin{tabular}{lccc}
  \toprule
  Horizon & \(d=64\) & \(\mathbf{d=128}\) & \(d=256\) \\
  \midrule
  10 & 0.193 & \textbf{0.191} & 0.195 \\
  20 & 0.267 & \textbf{0.264} & 0.270 \\
  30 & 0.425 & \textbf{0.423} & 0.430 \\
  40 & 0.655 & \textbf{0.650} & 0.660 \\
  50 & 0.895 & \textbf{0.893} & 0.900 \\
  \bottomrule
  \end{tabular}
\end{table}

\subsection{Number of Sampled Futures \(K\)}

Finally, we vary the number of generated trajectories \(K\in\{5,10,20,50\}\). Table~\ref{tab:sensitivity_k} shows that \(K=20\) achieves the best trade-off: fewer samples underfit the distribution, while more samples provide no further gain.

\begin{table}[h!]
  \footnotesize
  \centering
  \caption{Sensitivity to Number of Sampled Futures \(K\).}
  \label{tab:sensitivity_k}
  \begin{tabular}{lcccc}
  \toprule
  Horizon & \(K=5\) & \(K=10\) & \(\mathbf{K=20}\) & \(K=50\) \\
  \midrule
  10 & 0.198 & 0.192 & \textbf{0.191} & 0.191 \\
  20 & 0.275 & 0.266 & \textbf{0.264} & 0.264 \\
  30 & 0.437 & 0.425 & \textbf{0.423} & 0.424 \\
  40 & 0.672 & 0.654 & \textbf{0.650} & 0.650 \\
  50 & 0.989 & 0.898 & \textbf{0.893} & 0.894 \\
  \bottomrule
  \end{tabular}
\end{table}

\bigskip
Across all five studies, RHINO’s default settings—three scales, moderate hyperedge sizes, a single latent MLP layer, embedding dimension \(128\), and \(K=20\) samples—reside at or near the minima of the RMSE curves. Statistical trends indicate that medium-sized group interactions and compact latent encoders capture the essential dynamics of highway traffic without overfitting or undue complexity.

\section{Baseline Comparison.}
We include two contemporary predictors—\emph{GameFormer}~\cite{huang2023gameformer} and \emph{MultiPath++}~\cite{varadarajan2022multipath++}—as complementary points of reference to our main comparison. GameFormer frames multi-agent forecasting through a game-theoretic lens with intention-aware policy reasoning, while MultiPath++ is an anchor-based, multi-modal regressor with strong near-term, lane-aligned performance.

\begin{figure}[!ht]
    \centering
    \includegraphics[width=0.8\textwidth]{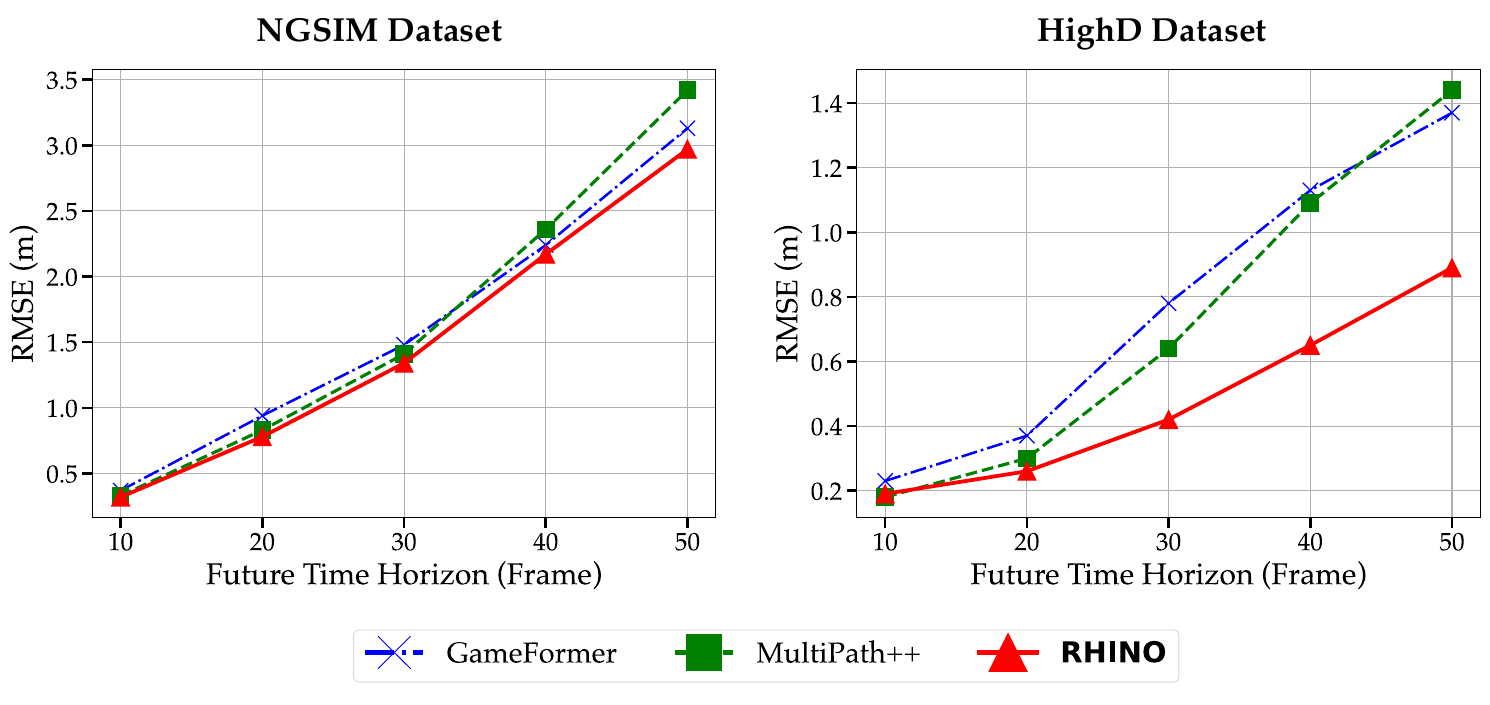}
    \caption{Prediction error obtained by new baseline models in RMSE on NGSIM dataset (left) and HighD dataset (right).}
    \label{fig_result_comp_new}
    \end{figure}

As summarized in Table~\ref{tab:comparison_new} and Figure~\ref{fig_result_comp_new}, \texttt{RHINO} achieves the best RMSE on NGSIM across all horizons, and on HighD it surpasses both baselines from 20 to 50 frames, with gaps increasing at longer horizons. MultiPath++ exhibits a slight advantage at the shortest (10-frame) horizon on HighD, consistent with its design focus on near-term mode selection. Overall, these results reinforce our central claim: explicit, hypergraph-based group-wise reasoning confers systematic benefits as the prediction horizon grows.

\begin{table}[!ht]
        \footnotesize
        \caption{Prediction Error Obtained by New Baseline Models.}
        \label{tab:comparison_new}
        \centering
        \begin{tabular}{>{\centering\arraybackslash}m{1.8cm} >{\centering\arraybackslash}m{1.05cm}
                        >{\centering\arraybackslash}m{2.4cm} >{\centering\arraybackslash}m{2.4cm} >{\centering\arraybackslash}m{1.8cm}}
            \toprule
            Dataset & Horizon (Frame) & GameFormer~\cite{huang2023gameformer} & MultiPath++~\cite{varadarajan2022multipath++} & \texttt{RHINO} \\
            \midrule
            \multirow{5}{*}{NGSIM}
              & 10 & 0.37 & \underline{0.33} & \textbf{0.32} \\
              & 20 & 0.94 & \underline{0.83} & \textbf{0.78} \\
              & 30 & 1.48 & \underline{1.41} & \textbf{1.34} \\
              & 40 & \underline{2.24} & 2.36 & \textbf{2.17} \\
              & 50 & \underline{3.13} & 3.42 & \textbf{2.97} \\
            \midrule
            \multirow{5}{*}{HighD}
              & 10 & 0.23 & \textbf{0.18} & \underline{0.19} \\
              & 20 & 0.37 & \underline{0.30} & \textbf{0.26} \\
              & 30 & 0.78 & \underline{0.64} & \textbf{0.42} \\
              & 40 & 1.13 & \underline{1.09} & \textbf{0.65} \\
              & 50 & \underline{1.37} & 1.44 & \textbf{0.89} \\
            \bottomrule
        \end{tabular}
    \end{table}

\section{Behavior‐Mode Enumeration} \label{app_behavior_mode}

For completeness, we summarize both the full nine joint modes and the three‐mode lateral simplification used in GIRAFFE and RHINO.

\subsection{Joint Modes}
In our prior work, each vehicle’s behavior was decomposed into both a lateral and a longitudinal component, yielding
\begin{equation}
\begin{aligned}
\mathcal{M}_{\rm lat}
&= \bigl\{\,L\ (\text{Left lane change}),\;
           K\ (\text{Lane keeping}),\;
           R\ (\text{Right lane change})\bigr\},\\
\mathcal{M}_{\rm lon}
&= \bigl\{\,A\ (\text{Acceleration}),\;
           C\ (\text{Constant speed}),\;
           D\ (\text{Deceleration})\bigr\},\\
\mathcal{M}
&= \mathcal{M}_{\rm lat}\times \mathcal{M}_{\rm lon}\;=\; 
   \{(m_{\rm lat},m_{\rm lon})\mid m_{\rm lat}\in \mathcal{M}_{\rm lat},\,m_{\rm lon}\in \mathcal{M}_{\rm lon}\},
\end{aligned}
\end{equation}
so that
\begin{equation}
|\mathcal{M}| \;=\; 3\times3 \;=\; 9,
\end{equation}
with the nine joint modes explicitly:
\begin{equation}
\{(L,A),\,(L,C),\,(L,D),\,(K,A),\,(K,C),\,(K,D),\,(R,A),\,(R,C),\,(R,D)\}.
\end{equation}
Each behavior node \(v^b_{i,(m_{\rm lat},m_{\rm lon})}\) in the Agent–Behavior Graph represents one of these mode combinations.

\subsection{Lateral‐Only Simplification}
In RHINO, to focus on the dominant group interactions on multi-lane highways, we restrict attention to the three lateral maneuvers:
\begin{equation}
\mathcal{M}_{\rm lat}
= \bigl\{\,L\ (\text{Left lane change}),\;
           K\ (\text{Lane keeping}),\;
           R\ (\text{Right lane change})\bigr\},
\quad
M = |\mathcal{M}_{\rm lat}| = 3.
\end{equation}
Consequently, each agent \(i\) is expanded into exactly three behavior nodes:
\begin{equation}
\{\,v^b_{i,L},\;v^b_{i,K},\;v^b_{i,R}\}.
\end{equation}
Longitudinal actions—Acceleration, Constant speed, Deceleration—are modeled implicitly within the CVAE latent space and the GRU decoder, preserving speed‐related uncertainty without increasing the hypergraph’s node count.

\medskip
\noindent\textbf{Feasibility masking.}  
At each time step, we apply a binary mask to invalidate any lateral mode that is physically impossible (e.g.\ a left-lane change \(L\) when already in the leftmost lane).  The logit for any masked mode is set to \(-\infty\) prior to softmax, ensuring only valid maneuvers appear as nodes in \(\mathcal{V}^b\).

\section{Diffusion Graph Convolutional Networks (DGCN)} \label{app_dgcn}
The Diffusion Graph Convolutional Networks (DGCN) module models bidirectional dependencies between nodes in the graph embedding, following \cite{wu2021inductive, li2017diffusion}. The DGCN layer, denoted $DGCN(\cdot)$, applies diffusion convolution to the graph signal in both forward and reverse directions:
\begin{equation}
    \begin{split}
        H_{l+1}
    &= DGCN(H_l) \\
    &= \sum_{k=1}^{K} \left(T_k(\bar A_f) \cdot \tilde H_l \cdot \Theta_{f,l}^k + T_k(\bar A_b) \cdot \tilde H_l \cdot \Theta_{b,l}^k \right)
    \end{split}
\end{equation}
In this equation, $\tilde H_{l+1}$ represents the output of the $l$-th layer, with the input to the first layer being the masked feature matrix $X$. The forward transition matrix $\bar A_f = A/\text{rowsum}(A)$ captures downstream node dependencies, while the backward transition matrix $\bar A_b = A^T/\text{rowsum}(A^T)$ captures upstream dependencies. The function $T_k(\cdot)$ represents a Chebyshev polynomial used to approximate convolution with the $k$-th order neighbors of each node. This is expressed as $T_k(X) = 2X \cdot T_{k-1}(X) - T_{k-2}(X)$. Learnable parameters $\Theta_{b,l}^k$ and $\Theta_{f,l}^k$ assign weights to input data in the $l$-th layer. The forward diffusion process captures the influence of surrounding vehicles on a given vehicle, while the reverse diffusion process models the influence that the vehicle transmits to its surroundings.

\section{Gumbel Softmax Distribution} \label{app_gumbel}
The Gumbel-Softmax distribution \cite{jang2016categorical} provides a differentiable approximation of the categorical distribution, crucial for neural networks that rely on gradient-based optimization. Traditional categorical sampling, involving non-differentiable operations like argmax, poses challenges for backpropagation. Gumbel-Softmax addresses this by replacing argmax with a differentiable softmax function, allowing gradients to flow during training. Based on the Gumbel-Max trick, which adds Gumbel noise to log-probabilities followed by argmax, the Gumbel-Softmax substitutes softmax to maintain differentiability. The formula of the sample vectors $d$ is:

\begin{equation}
    d_i = \frac{\exp \left( 
    \left( \log (\pi_i) + g_i \right)/\tau\right)}{\sum_{j=1}^{k} \exp\left( \left(\log(\pi_j) + g_j\right)/\tau \right)}, \text{ for } i=1,\dots,k
\end{equation}
Here, $\pi_i$ are categorical probabilities, $g_i$ is Gumbel noise, and $\tau$ controls the distribution’s sharpness. As $\tau$ approaches zero, the Gumbel-Softmax approximates a one-hot distribution. Larger $\tau$ values result in a smoother distribution, useful for early training exploration. This property makes Gumbel-Softmax valuable for tasks like variational autoencoders and reinforcement learning, where discrete sampling with differentiability is needed.

\section{Computational Efficiency and Real‐Time Performance}
\label{app:compute}

\noindent\textbf{Hardware and Software.}
All experiments were carried out on a desktop workstation with an NVIDIA RTX 4090 GPU and an Intel Core i9-14900K CPU.  Our implementation uses PyTorch 2.4, and CUDA 12.6, reflecting a modern deep‐learning environment.

\medskip
\noindent\textbf{Training Protocol.}
RHINO is initialized from a GIRAFFE model pretrained on the same highway trajectory data.  We fine‐tune \texttt{RHINO} on the HighD training set, comprising 436 702 trajectory windows, with a batch size of 64.  Optimization is performed using Adam (learning rate \(1\times10^{-3}\)).  Across 50 epochs, the total wall‐clock time is 403 minutes, averaging 8.06 minutes per epoch ((approximately 7.55 s per batch).

\medskip
\noindent\textbf{Inference Benchmarks.}
We evaluate \texttt{RHINO} on the HighD test set (187 143 windows), sampling \(K=20\) futures per input.  
    In batched mode (batch size = 64), \texttt{RHINO} processes the entire set in 200.5 seconds, equivalent to 68.6 ms per batch or 1.07 ms per window.  
    For real-time deployment simulation (batch size = 1), we measure 20.4 ms per window. This latency meets stringent on‐vehicle requirements for motion prediction in autonomous driving~\cite{mandalapu2025real,wang2024optimizing}.

\medskip
We summarize these results in Table \ref{tab:compute_summary}.
\begin{table}[h!]
  \footnotesize
  \centering
  \caption{RHINO Computational Performance Summary}
  \label{tab:compute_summary}
  \begin{tabular}{lcccccc}
    \toprule
    \textbf{Stage} 
      & \textbf{Data Windows} 
      & \textbf{Batch Size} 
      & \textbf{Total Time} 
      & \textbf{Latency (ms/batch)} 
      & \textbf{Latency (ms/sample)} \\
    \midrule
    Training  
      & 436\,702  
      & 64  
      & 403 min  
      & 70.9  
      & 1.11 \\
    Inference (batched)  
      & 187\,143  
      & 64  
      & 200.5 s  
      & 68.6  
      & 1.07 \\
    Inference (single)  
      & 187\,143  
      & 1  
      & 3,823 ms  
      & --   
      & 20.4 \\
    \bottomrule
  \end{tabular}
\end{table}